\documentclass[10pt,twocolumn,letterpaper]{article}

\usepackage[accsupp]{axessibility}

\usepackage[pagenumbers]{iccv} 

\usepackage{colortbl}
\usepackage{ulem}
\usepackage{multirow}

\newcommand{\ourdataset}{RPINE\xspace}
\newcommand{\ours}{TMR\xspace}

\newcommand{\smallbreakparagraph}[1]{\smallbreak \noindent \textbf{#1}}

\definecolor{iccvblue}{rgb}{0.21,0.49,0.74}
\usepackage[pagebackref,breaklinks,colorlinks,allcolors=iccvblue]{hyperref}

\def\paperID{2220} 
\def\confName{ICCV}
\def\confYear{2025}

\title{
Few-Shot Pattern Detection via Template Matching and Regression  
}

\author{
Eunchan Jo\qquad Dahyun Kang\qquad Sanghyun Kim\qquad Yunseon Choi\qquad Minsu Cho \vspace{1.5mm}\\
Pohang University of Science and Technology (POSTECH), South Korea \\
{\tt\small \href{https://cvlab.postech.ac.kr/research/TMR}{https://cvlab.postech.ac.kr/research/TMR}}
}

\usepackage{amsmath,amsfonts,bm}

\def\eqref#1{equation~\ref{#1}}

\def\1{\bm{1}}

\def\mB{{\bm{B}}}

\def\mF{{\bm{F}}}

\def\mI{{\bm{I}}}

\def\mP{{\bm{P}}}

\def\mT{{\bm{T}}}

\DeclareMathAlphabet{\mathsfit}{\encodingdefault}{\sfdefault}{m}{sl}
\SetMathAlphabet{\mathsfit}{bold}{\encodingdefault}{\sfdefault}{bx}{n}

\begin{document}
\maketitle

\begin{abstract}
We address the problem of few-shot pattern detection, which aims to detect all instances of a given pattern, typically represented by a few exemplars, from an input image.
Although similar problems have been studied in few-shot object counting and detection (FSCD), previous methods and their benchmarks have narrowed patterns of interest to object categories and often fail to localize non-object patterns. 
In this work, we propose a simple yet effective detector based on template matching and regression, dubbed \ours.
While previous FSCD methods typically represent target exemplars as spatially collapsed prototypes and lose structural information, we revisit classic template matching and regression.
It effectively preserves and leverages the spatial layout of exemplars through a minimalistic structure with a small number of learnable convolutional or projection layers on top of a frozen backbone.
We also introduce a new dataset, dubbed RPINE, which covers a wider range of patterns than existing object-centric datasets.
Our method outperforms the state-of-the-art methods on the three benchmarks, RPINE, FSCD-147, and FSCD-LVIS, and demonstrates strong generalization in cross-dataset evaluation.
\end{abstract}    

\vspace{-2mm}
\section{Introduction}
\label{sec:intro}

Few-shot detection aims to identify target patterns with minimal labeled examples.
While significant progress has been made in few-shot object detection~\cite{fan2020few, metafasterrcnn, xiao2022few, yan2019meta, fan2021generalized}, most existing methods remain object-centric, focusing primarily on identifying object-level patterns with relatively clear boundaries. 
However, many real-world applications require detecting arbitrary target patterns that extend beyond objects to include structural, geometric, or abstract patterns across diverse visual data. 
Despite recent progress based on deep neural networks, current methods still fall short in addressing these broader pattern detection tasks.
Furthermore, the object-centric design of conventional few-shot detectors may lead to performance degradation when the target object lacks clear boundaries or when occlusion and deformation cause its boundaries to become indistinct.

\begin{figure}[t!]
\vspace{-3mm}
	\centering
	\small
        \includegraphics[width=0.99\linewidth]{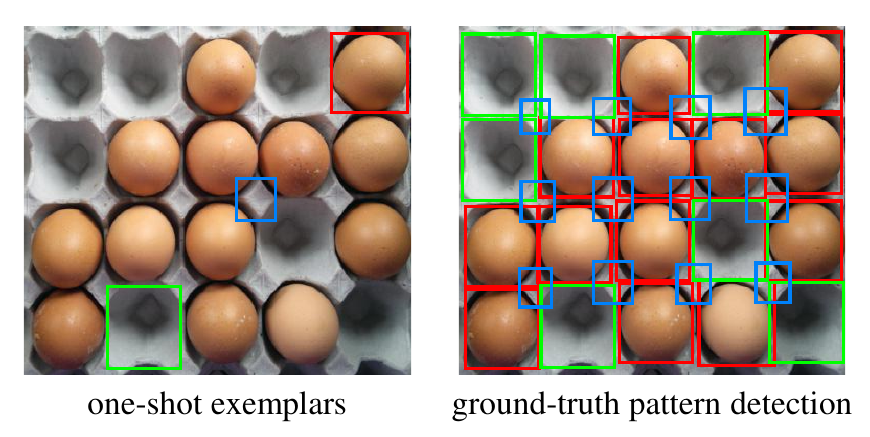}     
        \vspace{-3mm}
 \caption{Few-shot pattern detection.
 Given a few exemplar for each target pattern (left), the task is to detect all matching instances of each pattern (right). This example include non-object patterns (\eg, green and blue) as well as object patterns (\eg, red).
 }
 \label{fig:teaser}
 \vspace{-4mm}
\end{figure}

\begin{figure}[t!]
	\centering
	\small
    \centering
    \includegraphics[width=0.99\linewidth]{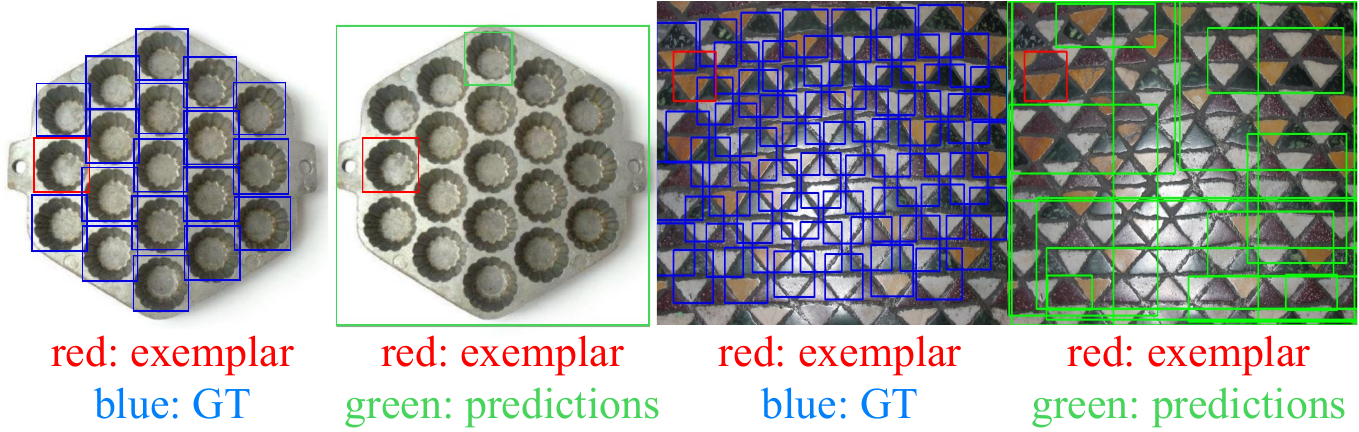}
    \vspace{-2mm}
 \caption{Few-shot detectors trained with strong object prior are often biased to objects instances and struggle to detect non-objects. 
 }    \label{fig:fsod_failures}

\vspace{-5mm}
\end{figure}

The task of few-shot pattern detection is illustrated in Fig.~\ref{fig:teaser}.
Recent related research topics for few-shot detection include few-shot counting and detection~\cite{cdeter}, and few-shot object detection~\cite{fan2020few}.
Both aim to reduce the annotation cost of object detection~\cite{yolo, fastrcnn, fcos, fpn} by learning to detect all instances of given support exemplars. 
Consequently, many of these methods are heavily biased on object-centric benchmarks prior~\cite{pascal, coco, lvis, cdeter}.
In addition, many recent approaches~\cite{geco, pseco, xiao2022few,fan2020few,metafasterrcnn,yan2019meta} represent the support exemplars as the spatially pooled vector, often named as a prototype~\cite{protonet}.
While this pooling strategy is effective for detecting objects, it collapses the geometric properties, such as the shape and structure of the support exemplars.
As a consequence, these methods tend to underperform when detecting non-object geometric patterns such as object parts or shape-intensive elements as shown in  Fig.~\ref{fig:fsod_failures}.

\begin{figure*}[t!]
    \vspace{-5mm}
	\centering
	\small
    \includegraphics[width=0.99\linewidth]{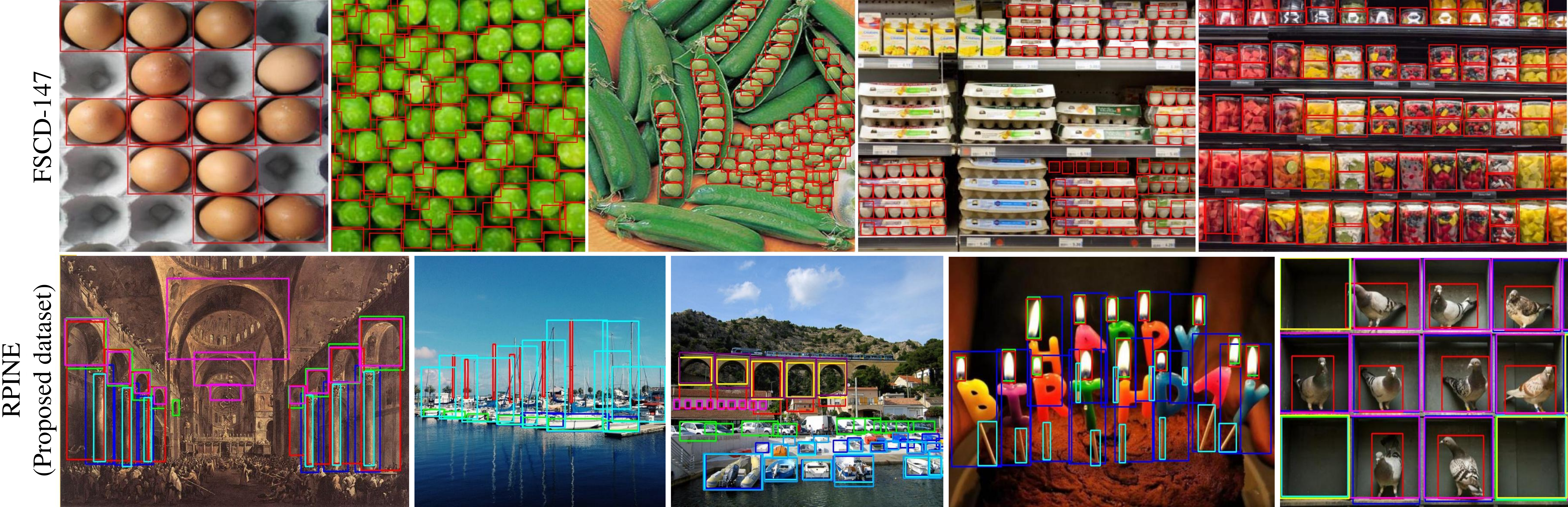}
    \vspace{-2mm}
 \caption{Annotation examples of FSCD-147~\cite{cdeter} and \ourdataset.
 FSCD-147 is annotated with the repetitive object-level patterns but disregards the repetition of non-object patterns such as the egg tray pattern under the eggs in the first image.
 \ourdataset is annotated with arbitrarily noticed repeated patterns which include non-object patterns and nameless parts of objects.
 Plus, FSCD-147 is annotated with a single pattern, while \ourdataset is annotated with all existing repetitions (marked with different colors) recognized by three different annotators.}
\label{fig:dataset}
\vspace{-5mm}
\end{figure*}

In this work, we revisit the classic template matching strategy and propose a simple yet effective few-shot detector for arbitrary patterns.
The proposed method, dubbed template matching and regression (TMR), is designed to be aware of the structure and shape of given exemplars.
Given an input image, TMR first extracts a feature map using a backbone network.  
It then crops a template feature from the support exemplar’s bounding box using a template extraction technique based on RoIAlign~\cite{maskrcnn}.
This template is correlated with the image feature map to produce a template matching feature map.
Using this correlation map, the model learns bounding box regression parameters to rectify the support exemplar's box size adaptively.
This process, termed support-conditioned regression, enables the model to handle support exemplars of varying sizes more effectively.
Notably, \ours consists only of a few $3\times3$ and linear projections without any complicated modules such as cross-attention, commonly used in prior work~\cite{cdeter, geco}.

Although \ours is designed for general pattern detection, existing benchmarks (\eg, FSCD-147~\cite{cdeter}, FSCD-LVIS ~\cite{cdeter}) mainly target object-level patterns, limiting comprehensive evaluation.
To address this, we introduce a new dataset, Repeated Patterns IN Everywhere (\ourdataset), which covers diverse repeated patterns in the real world.
\ourdataset contains images with varying degrees of objectness, from well-defined object-level patterns to non-object patterns, all annotated with bounding boxes via crowd-sourcing.
Compared to FSCD datasets, \ourdataset provides broader coverage, including both non-object patterns and nameless parts of objects, as illustrated in Fig.~\ref{fig:dataset}.

\ours demonstrates strong performance in detecting repeated patterns, not only on \ourdataset but also on the FSCD benchmarks, FSCD-147 and FSCD-LVIS~\cite{cdeter}.  
In particular, \ours is especially effective on \ourdataset, includes diverse patterns with minimal object priors.
Notably, our simple architecture contributes to improved generalization across datasets.
Our contribution is summarized as follows:
\begin{itemize}
    \item We generalize the few-shot object counting and detection to a pattern detection task that does not assume objectness in either the target patterns or exemplars.
    \item We present a simple yet effective pattern detector by refining template matching, which efficiently detects coherent patterns guided by exemplars.
    \item We introduce a new densely annotated dataset, \ourdataset, which covers diverse repetitive patterns in the real world, ranging from object-level patterns to non-object patterns.
    \item \ours not only outperforms the state-of-the-art FSCD models on \ourdataset and FSCD-LVIS but also achieves strong cross-dataset generalization.
\end{itemize}


\begin{figure*}[t!]
\vspace{-6mm}
	\centering
	\small
    \includegraphics[width=0.99\linewidth]{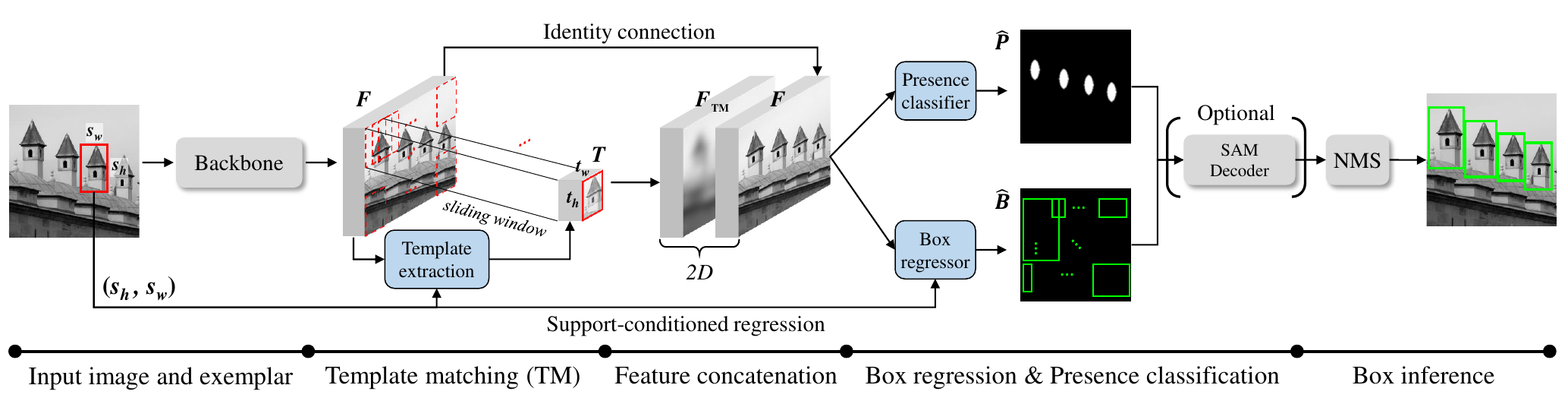}
 \vspace{-3mm}
 \caption{\textbf{Template Matching and Regression (TMR).} A template feature map is cropped from the image feature map and then correlated with the image feature map via channel-wise template matching.
 The TM feature map and the image feature map are concatenated.
 For each feature map point, the box regressor predicts the shifting and scaling parameters of the support exemplar's size, and the presence classifier scores the presence map.
 Both of them consist of a $3\times3$ convolution and a linear projection without any complicated layers.
 }
\label{fig:method}
\vspace{-4mm}
\end{figure*}

\section{Related work}
\label{sec:related_work}
\smallbreakparagraph{Few-shot object detection~(FSOD)} aims to detect objects of novel classes using only a few support images of novel classes.
Existing methods can be roughly categorized into two groups: finetuning based~\cite{xiao2022few,guirguis2023niff,sun2021fsce,fan2021generalized,wang2020frustratingly} and meta-learning based~\cite{kang2019few,fan2020few,han2022few,han2021query,metafasterrcnn,yan2019meta, zhang2023detect} approaches.
Despite significant progress, finetuning methods require retraining for every new classes.
For pattern detection, it is not desired to fine-tune each arbitrary pattern with indistinct object categorization.
In contrast, meta-learning methods avoid class-specific fine-tuning and typically construct prototypes from support images for classifying the bounding boxes of the query image.
These methods often apply global average pooling to obtain prototypes from each support exemplar, which collapses the spatial structure of the exemplar.
While such spatially collapsed prototypes may effectively capture object-level bounding boxes, they struggle with non-object patterns that exhibit complex spatial structures.

\smallbreakparagraph{Few-shot counting~(FSC)} aims to count objects given a few exemplars (typically 1 to 3) within an image.
Previous methods~\cite{lu2019class,ranjan2022vicinal,lin2021object,ranjan2021learning,you2022iterative} have tackled this task using density-map regression without detecting the individual bounding boxes.
Towards localization-based counting, few-shot counting and detection (FSCD) is proposed to combine FSC and detection.
Such models include Counting-DETR~\cite{cdeter}, SAM-C~\cite{ma2023samcountanythingempirical}, DAVE~\cite{pelhan2024dave}, PseCo~\cite{pseco}, and GeCo~\cite{geco}.
However, similar to the FSOD methods, these models typically generate a prototype by spatially averaging exemplar bounding boxes.
This discards the spatial structure of the exemplars, potentially losing important cues and details for accurate counting and detection.
In contrast, our pattern-detection method preserves the spatial structure of the exemplar for pattern matching.

\smallbreakparagraph{Template matching.}
Template matching~\cite{marola1989using, korman2013fast, tsai2002rotation,mohan2001example, hays2006discovering, turina2001efficient, matas2004robust} has been widely used from the beginning of computer vision and pattern recognition and also adopted in convolution-based neural detectors with additional regression~\cite{fpn, fastrcnn}. 
Given a 2D template, template matching identifies the matching region by sliding-window fashion.

\smallbreakparagraph{Repetitive pattern detection.}
Detecting repetitive patterns is trivial for humans but remains challenging in computer vision.
Early research focused on geometrically constrained settings where patterns are nearly regular and aligned~\cite{grunbaum1987tilings, lin1997extracting, han2008regular, pauly2008discovering}.
Based on this near-regularity, early methods assumed a global repetitive structure and discovered the repetition~\cite{liu2004near, hays2006discovering, park2009deformed, liu2015patchmatch, lin1997extracting} in a top-down manner.
In contrast, later methods~\cite{lin2006tracking, gao2009unsupervised, spinello2010exploiting} define the smallest repeating unit, called a texel~\cite{leung1996detecting, cai2013translation, liu2013grasp} and identify all matching subparts based on the identified texels.
In our context, the texel corresponds to the given exemplar.


\section{Few-shot pattern detection}
\label{sec:problem}

Given an input image $\mI\in \mathbb{R}^{H_0 \times W_0 \times 3}$, the goal of few-shot pattern detection is to predict matching patterns with a given set of support exemplars.
With an abuse of notations, the model is given the set of support exemplar, $\mathcal{E} = \{e_1, \cdots, e_{N_s}\}$ and aims to predict the corresponding ground-truth bounding boxes of pattern $\mathcal{B} = \{b_1,\cdots,b_{N_{p}}\}$, where $N_p$ and $N_s$ denotes the number of ground-truth bounding boxes of the pattern and the number of exemplars (typically referred to as the ``shot''), respectively.
The input exemplars and the output patterns are both represented as bounding boxes parameterized by their center coordinates, width, and height: $b_{i}, e_{i} \in \mathbb{R}^{4}$.


\section{Template matching and regression (TMR)}
\label{sec:method}
For clarity, we primarily focus on the one-shot, single-scale pattern detection setting, where a single support exemplar is given and a single-resolution image feature map is used.
However, \ours can be effortlessly extended to few-shot and multi-scale scenarios, as described in Sec.\ref{sec:inference} and Sec.\ref{sec:multi-scale}.

The overall architecture of \ours is illustrated in Fig.~\ref{fig:method}.
The input image $\mI$ is first encoded by a backbone such as ViT~\cite{sam} to extract a feature map $\mF \in \mathbb{R}^{H \times W \times D}$.
The template feature $\mT \in \mathbb{R}^{t_h \times t_w \times D}$ is then obtained through the template extraction process, which uses RoIAlign~\cite{maskrcnn} to crop a region from the image feature map $\mF$ based on the support exemplar's size ($s_h, s_w$).
In RoIAlign, unlike previous fixed-size pooling methods~\cite{cdeter, geco}, the model adaptively determines the size of $\mT$ to fit the corresponding size on $\mF$ by rounding up the size of the  exemplars on $\mF$ as described in Sec.~\ref{sec:supp_template_extraction}.
This preserves the spatial alignment between $\mT$ and $\mF$ with translation.
The image feature map $\mF$ and template feature $\mT$ are correlated by template matching (Sec.~\ref{sec:tm}), which outputs the template-matching feature $\mF_{\text{TM}}$.
The concatenation of the feature maps $\mF$, $\mF_{\text{TM}}$ is fed to the subsequent box prediction module, which consists of a pattern box regressor and a pattern presence classifier.
The pattern box regressor predicts the localization bounding boxes: $\hat{\mB}$, which is parameterized by scaling and shifting factors of the given exemplar's size. 
The pattern presence classifier predicts the presence score of the pattern: $\hat{\mP}$.
A box proposal is then generated on each feature map point based on the combination of $\hat{\mB}$ and $\hat{\mP}$ (Sec.~\ref{sec:tr}).
At inference, bounding boxes with low presence scores are removed by Non-Maximum Suppression (NMS).
SAM decoder~\cite{sam} can be optionally applied for box refinement before NMS.

\subsection{Template matching (TM)}
\label{sec:tm}
We are motivated to detect arbitrary patterns given a support exemplar of either an object or not.
Non-object patterns often lack high-level semantics yet exhibit low-level structural features, thus, preserving the spatial structure of the support exemplar is crucial.
Inspired by traditional template matching~\cite{marola1989using, korman2013fast, tsai2002rotation}, we compute the matching score between the image and the support exemplar to detect the locations of the pattern.
Specifically, template matching cross-correlates the feature map $\mF$ and the template feature $\mT$ by centering $\mT$ at each $(x, y)$ position in $\mF$.
The resultant template-matching (TM) feature $\mF_{\text{TM}}$ is obtained as:
\begin{equation}
    \mF_{\text{TM}}(x, y) = \frac{1}{t_wt_h}\sum_{x',y'}\! \mF(x + x'\! -\! \lfloor\!\frac{t_w}{2}\!\rfloor, y + y'\! -\lfloor\!\frac{t_h}{2}\!\rfloor)\mT(x'\!, y'), \label{eq:tm}
\end{equation}
where $\lfloor \cdot \rfloor$ denotes the floor operation used for centering $\mT$ at each $(x,y)$,
and $(x', y')$ ranges within the template coordinates: $\in [0, t_w) \times [0, t_h)$.
Note that Eq.~\ref{eq:tm} computes the correlation by channel-wise multiplication, resulting in $\mF_{\text{TM}}\in\mathbb{R}^{H \times W \times D}$.
Finally, the TM feature $\mF_{\text{TM}}$ is concatenated with the feature map $\mF$: $\mF_{\text{P}} = [\mF_\text{TM};\mF]\in\mathbb{R}^{H \times W \times 2D}$.
The concatenated feature map $\mF_{\text{P}}$ is fed to the subsequent box prediction module.

\subsection{Support-conditioned box regression}
\label{sec:tr}
Our box prediction module consists of \textit{a pattern box regressor} and \textit{a pattern presence classifier} following the architecture of an anchor-free detection methods~\cite{fcos}.

The pattern box regressor $g_{\text{B}}$ consists of a $3\times3$ convolutional layer followed by a linear layer and predicts the four localization parameters $(\Delta x, \Delta y, \alpha_w, \alpha_h)$ for each feature map point.
Unlike the methods~\cite{cdeter,geco} that directly regress absolute box parameters, our method performs \textit{support-conditioned} regression.
Bounding box parameters are predicted by scaling and shifting the support exemplar's size.
This helps to dynamically adjust the exemplar's size to predict a target box size.
A predicted bounding box at a feature point $(x, y)$ shifts and scales the support exemplar's size $(s_w, s_h)$ such as:
\begin{align}
(x + s_w\Delta x, y + s_h \Delta y, e^{\alpha_w} s_w, e^{\alpha_h} s_h). \label{eq:bbox_params}
\end{align}

The pattern presence classifier $g_{\text{P}}$ consists of a linear layer and predicts presence scores, which represent the confidence of the predicted bounding boxes at each feature map point.
The aforementioned procedure of the box regressor $g_{\text{B}}$ and the presence classifier $g_{\text{P}}$ is summarized as:
\vspace{-1mm}
\begin{align}
    \hat{\mB} &= g_{\text{B}}(\mF_\text{P}), \quad \hat{\mB} \in \mathbb{R}^{H \times W \times 4}, \\
    \hat{\mP} &= \sigma(g_{\text{P}}(\mF_\text{P})), \quad \hat{\mP}
\in \mathbb{R}^{H \times W \times 1}, \label{eq:boxhead}
\end{align}
\vspace{-1mm}
where $\sigma$ denotes the sigmoid function.

\subsection{Inference}
\label{sec:inference}
At inference, we first remove bounding boxes whose presence score is lower than a threshold $\tau$ to filter out low-confidence predictions.
Afterward, we can optionally apply box localization refinement via the SAM decoder~\cite{sam} as often adopted by the FSCD work~\cite{ma2023samcountanythingempirical, pseco, geco}.
Following \cite{geco,pseco}, we input the predicted box coordinates into the SAM prompt encoder to obtain the prompt feature.
The obtained prompt feature and the image feature extracted from the SAM backbone are fed to the SAM decoder, which further refines input box coordinates.
To obtain the final box prediction, $\hat{\mathcal{B}}$, we apply NMS on the bounding boxes.

\ours is easily extended for few-shot inference without re-training.
When multiple support exemplars are given, we perform the above process for each exemplar individually and then aggregate the results before applying NMS.


\begin{table*}[t!]
\centering
\small
\vspace{-5mm}
\scalebox{0.9}{
\tabcolsep=0.15cm
\begin{tabular}{l|lllc}
\toprule
    dataset & pattern tiling & repetition & object bias & multi-pattern annotations per image \\ \midrule
    Wallpaper~\cite{liu2013symmetry} & near regular \cite{liu2004near} & high & low & \\
    Pascal~\cite{pascal}, COCO~\cite{coco}  & - & low & high & \checkmark (multi-category)\\
    Wikiart~\cite{wikiart}, Frieze~\cite{frieze}, Counting bench~\cite{paiss2023teaching} & arbitrary & mid & high & \\
    FSCD-147~\cite{cdeter}, FSCD-LVIS~\cite{cdeter} & arbitrary & high & high  &\\
    \cellcolor{gray!20}\ourdataset  & \cellcolor{gray!20}arbitrary & \cellcolor{gray!20}high & \cellcolor{gray!20}low to high  & \cellcolor{gray!20}\checkmark \\
\bottomrule
\end{tabular}
}
\vspace{-2mm}
 \caption{
 Benchmark dataset comparison of related work. \ourdataset covers pattern or object repetitions in the real world. 
We classify `pattern-tiling' as \textit{near-regular} following~\cite{liu2004near}, \ie, elements with an almost periodic lattice with minor variations in shape, color or lighting, and as \textit{arbitrary} otherwise.
The `repetition' is classified based on the average number of repeated same class instances on an image: \textit{low} if $\leq$ 5, \textit{mid} if $\leq$ 20, and \textit{high} otherwise.
The `object-bias' is \textit{high} if the dataset annotations correspond to a predefined class set, and \textit{low} otherwise.
}
     \vspace{-4mm}
\label{tab:benchmark_comparison}
\end{table*}

\subsection{Learning objective} 
The training loss is composed of $\mathcal{L}_{\text{P}}$, which penalizes the presence score, and $\mathcal{L}_{\text{B}}$, which penalizes the bounding box regression. 
The presence loss $\mathcal{L}_{\text{P}}$ is the binary cross-entropy loss (BCELoss) between the predicted and the ground-truth center points of the pattern bounding boxes.
Instead of guiding the presence with a single pixel, we add a margin around each ground-truth center point and define the extended center point set as $\mathcal{X}_{\text{P}}$.
We detail this in Sec.~\ref{sec:supp_xp_details}.
The ground-truth presence label $\mP$ at a position $(x, y)$ is set to 1 if the position is in $\mathcal{X}_{\text{P}}$, and 0 otherwise.

The bounding box regression loss $\mathcal{L}_{\text{B}}$ is defined as the generalized IoU (gIoU) loss~\cite{rezatofighi2019generalized} between the predicted and the ground-truth bounding boxes.
This loss is defined on the positions where the ground-truth patterns exist.
For each center point of the pattern bounding box, the adjacent area of the center within the margin shares the corresponding bounding box parameter $\mB$.
The two loss functions are defined as the following:
\vspace{-1mm}
\begin{align}
    \mathcal{L}_{\text{P}} &= \sum_{(x,y)}^{\forall (x,y)} \text{BCELoss}(\hat{\mP}(x,y), \mP(x,y)), \\
    \mathcal{L}_{\text{B}} &= \sum_{(x,y)}^{(x,y) \in \mathcal{X}_{\text{P}}} \text{gIoULoss}(\hat{\mB}(x,y), \mB(x,y)).
\end{align}
\vspace{-1mm}
The overall training loss is the sum of them: $\mathcal{L} = \mathcal{L}_{\text{P}} + \mathcal{L}_{\text{B}}$.




\section{Proposed dataset: \ourdataset}
Existing benchmarks (\eg, FSCD-147~\cite{cdeter}, FSCD-LVIS ~\cite{cdeter}) focus on object-level patterns, limiting their use for general pattern detection.  
Therefore, we introduce a new pattern dataset, \ourdataset: Repeated Patterns In Everywhere.

\begin{figure}[t!]
	\centering
	\small
    \centering
\includegraphics[width=0.99\linewidth]{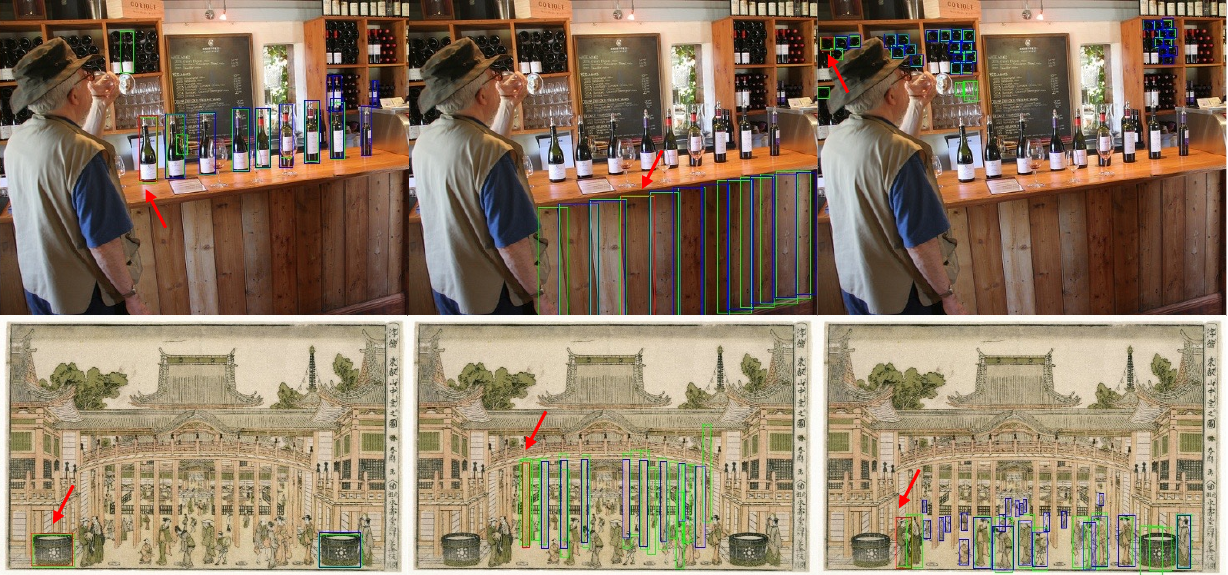}
\vspace{-2mm}
 \caption{GTs (blue) and \ours predictions (green) with different exemplars (red) from \ourdataset, which is the only dataset containing multiple patterns for each image among FSCD datasets. 
 }    \label{fig:different_exemplars}
     \vspace{-4mm}
\end{figure}

We collect images from the repetitive pattern detection literature and annotate them to contain various repetitive patterns in the wild.
Images are collected from FSC-147~\cite{cdeter}, FSCD-LVIS~\cite{cdeter}, Countbench~\cite{paiss2023teaching}, Wikiart~\cite{wikiart}, Frieze~\cite{mancini2018adding}, and Wallpaper~\cite{liu2013symmetry}.
Tab.~\ref{tab:benchmark_comparison} shows where \ourdataset stands among related datasets.
The dataset contains 4,362 images, divided into 3,925 and 435 for training and testing, respectively.
As shown in Fig.~\ref{fig:different_exemplars}, \ourdataset is annotated with multiple patterns per image.
\ourdataset is a suitable evaluation benchmark for multi-pattern detection within an image.

In the real world, a pattern cannot be rigorously defined.
We thus define the following criteria inspired by the definition of the translation symmetry~\cite{stenger2009timeless} to minimize the subjective variance among different annotators.
\begin{itemize}
    \item Number of patterns: max 3 different patterns are annotated per image if the image exhibits multiple patterns.
    \item Number of pattern instances: has no upperbound.
    \item Minimum Size: the width and height of a pattern instance must be at least 3\% of the shorter side of the image.
    \item Appearance variance: visually similar patterns with different scales/colors/semantics/rotation angles are annotated as the same pattern.
    \item Reflection: visually similar patterns but reflection symmetric patterns are annotated in different patterns.
    \item Occlusion: if visually similar patterns are occluded from each other, the visible parts are annotated.
\end{itemize}
We ask annotators to carefully draw bounding boxes on the recognized patterns by the above instructions as consistently as they can.
Despite of the instruction details, pattern are not pixel-perfectly defined across annotators. 
Therefore, we assign three individual annotators per image and include all the annotated patterns as ground truth.



\section{Experiments}
\label{sec:experiments}

\subsection{Dataset and metrics}
We evaluate our model on \ourdataset as well as on the two standard FSCD benchmarks: FSCD-147~\cite{cdeter} and FSCD-LVIS~\cite{cdeter}.
FSCD-147 contains a total of 6,135 images, with 3,659 for training, 1,286 for validation, and 1,190 for testing.
FSCD-LVIS seen-split contains a total of 6,195 images, with 4,000 for training, 1,181 for validation, and 1,014 for testing, covering 372 object categories. FSCD-LVIS unseen-split contains a total of 6,201 images, with 3,959 for training and 2,242 for testing where test-time object categories are not observed during training.

Following the evaluation protocol~\cite{cdeter,geco}, we report Mean Absolute Error (MAE) and Root Mean Squared Error (RMSE) for counting.
For detection, we report average precision with IoU thresholds of 0.5 and 0.75, denoted as AP50 and AP75, respectively, along with the averaged AP over IoU thresholds from 0.5 to 0.95 in increments of 0.05.

\subsection{Implementation details}
We use the pre-trained SAM-ViT/H~\cite{sam} of the patch size $16$ as the backbone and set it frozen during training, which returns 
the $64\times64\times256$ feature map.
Due to the input patchification encapsulated in the ViT backbones~\cite{vit}, we find the receptive field of the raw feature map too coarse to detect small instances.
We thus bilinearly interpolate the feature map resolution from $64\times64$ to $128\times128$ to produce a higher-resolution correlation map, which enables denser predictions and leads to better performance. (Tab.~\ref{tab:RPINE-multiscale})

The channel dimension is expanded to $512$ via a learnable linear projection.
\ours is trained with the learning rate of $10^{-4}$ with AdamW~\cite{loshchilov2019decoupledweightdecayregularization} and the batch size of 16 on four Nvidia RTX 3090 GPUs with 24 GB VRAM for 24 hours.
The box presence threshold $\tau$ for NMS set to 0.4, 0.3 for RPINE and FSCD-147, respectively.

\subsection{Comparison with state-of-the-art methods}
\begin{figure*}[t!]
	\centering
        \vspace{-4mm}
	\small
    \scalebox{0.85}{
    \includegraphics[width=0.99\linewidth]{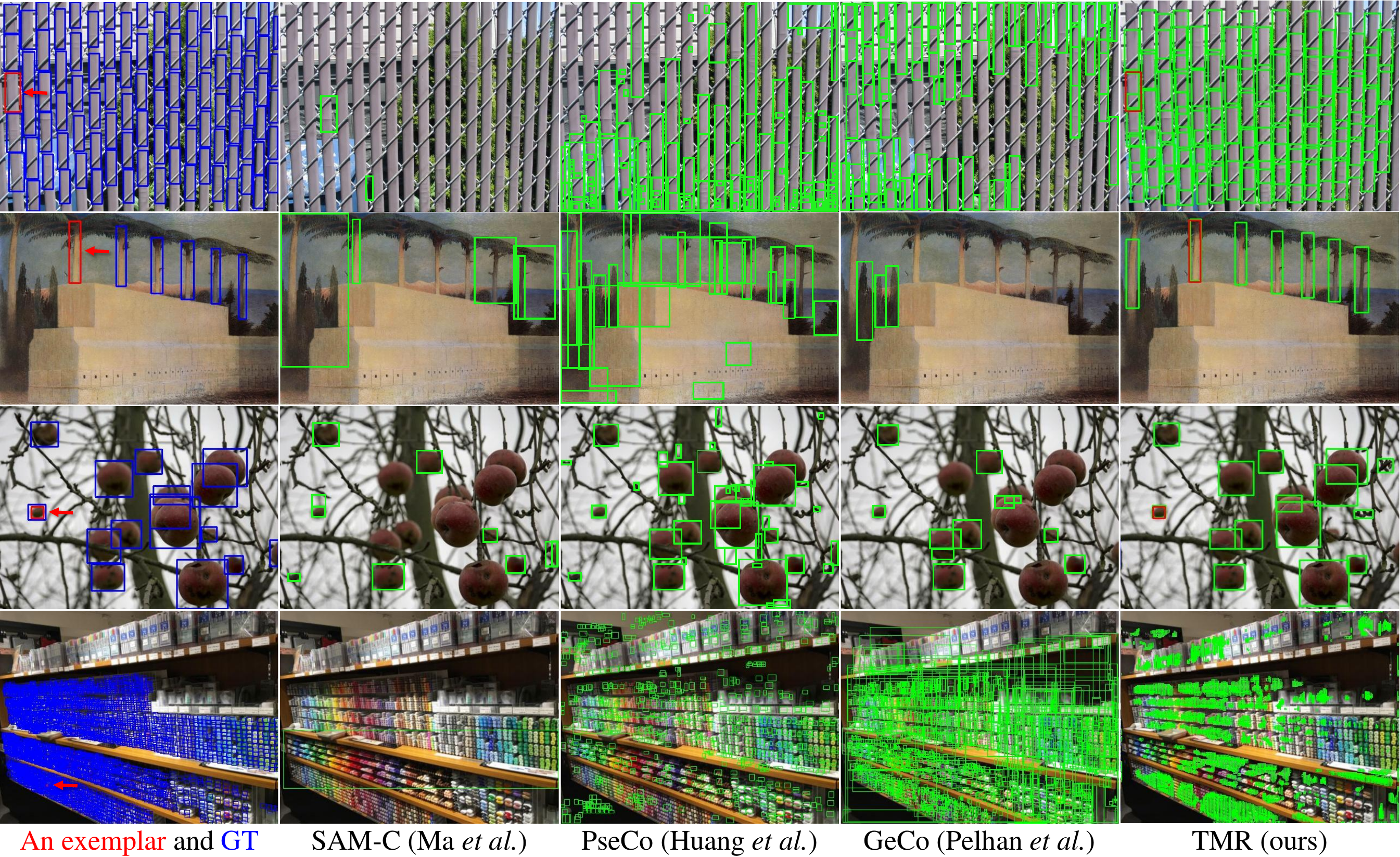}
    }
    \vspace{-2mm}
 \caption{Qualitative comparison with the state-of-the-art models on \ourdataset (the first two images) and FSCD-147 (the last two images).
 More visualization examples on three datasets are included in the supplementary materials Figs~\ref{fig:sup_qual_lvis}-\ref{fig:sup_qual_rpine}.
 }
\label{fig:qual}
\vspace{-4mm}
\end{figure*}

\smallbreakparagraph{RPINE.}
To demonstrate the effectiveness of our method in counting and detecting non-object patterns, we evaluate it on RPINE.
In Tab.~\ref{tab:RPINE-1shot-results}, \ours surpasses the previous FSCD methods with a large margin.
As the previous FSCD methods rely on prototypes for matching, they tend to struggle with non-object patterns that require an understanding of spatial details rather than semantics.
In contrast, \ours effectively detects non-object patterns by incorporating spatial details in the template matching process.
Note that \ourdataset is the only dataset equipped with multiple pattern annotations for each image in FSCD.
Figure~\ref{fig:different_exemplars} demonstrates the bounding box predictions with different exemplars, where \ours predicts bounding boxes adaptively to the given exemplars.
Figure~\ref{fig:qual} shows the qualitative comparisons with other FSCD methods where \ours accurately localize target patterns.

\smallbreakparagraph{FSCD-LVIS and FSCD-147.}
We compare \ours with existing methods that are dedicated to FSCD under the FSCD setting on FSCD-LVIS and FSCD-147.
The results in Tab.~\ref{tab:fscd-LVIS-results} demonstrate that \ours significantly outperforms previous state-of-the-art approaches.
Notably, when evaluated on the unseen split, where test-time object categories are not observed during training, \ours surpasses prior methods.
This suggests that \ours is less biased to object semantics during training potentially because \ours leverages the exemplar's spatial structure that generalizes across different categories.
In addition to FSCD-LVIS, Tab.~\ref{tab:fsc147-results} compares methods on FSCD-147, where \ours performs on par with the previous methods on both one-shot and three-shot settings.

\begin{table}[t!]
\small
\centering
\scalebox{0.87}{
\tabcolsep=0.15cm
\begin{tabular}{lcccccc}
\toprule
Method   & SD & MAE($\downarrow$) & RMSE($\downarrow$) & AP($\uparrow$) & AP50($\uparrow$) & AP75($\uparrow$) \\ \midrule
C-DETR~\cite{cdeter}           & & 9.58 & 21.24 & 13.88 & 32.20 & 10.22 \\
SAM-C~\cite{ma2023samcountanythingempirical}             & \checkmark & 18.77 & 37.14 & 18.80 & 34.04 & 18.74 \\
PseCo~\cite{pseco}             & \checkmark & 48.20 & 88.16 & 23.18 & 44.54 & 21.24 \\
GeCo~\cite{geco}              & \checkmark & 9.57 & 17.07 & 23.33 & 45.93 & 21.19 \\
\cellcolor{gray!20}\ours$_{\text{(ours)}}$     & \cellcolor{gray!20} & \cellcolor{gray!20}\textbf{8.45} & \cellcolor{gray!20}19.87 & 
\cellcolor{gray!20}\textbf{33.59} & \cellcolor{gray!20}\textbf{64.05} &  \cellcolor{gray!20}\textbf{30.52} \\
\cellcolor{gray!20}\ours$_{\text{(ours)}}$      & \cellcolor{gray!20}\checkmark & \cellcolor{gray!20}\textbf{8.30} & \cellcolor{gray!20}19.40 & \cellcolor{gray!20}\textbf{29.66} & \cellcolor{gray!20}\textbf{58.94} & \cellcolor{gray!20}\textbf{25.41} \\
\bottomrule
\end{tabular}
}
\vspace{-2mm}
\caption{
One-shot pattern counting and detection results on the RPINE dataset. SD denotes box refinement with the SAM decoder. All the models are trained by the official code. 
}
\label{tab:RPINE-1shot-results}
\vspace{-4mm}
\end{table}

\subsection{Analyses and ablation study}
\label{sec:ablation}
\smallbreakparagraph{\ours learns with less semantic object bias and generalizes well across datasets.}
We compare \ours and GeCo~\cite{geco} in the cross-dataset scenarios by evaluating the trained models on different datasets that are unseen during training.
As shown in Tab.~\ref{tab:cross-dataset}, \textit{\ours presents overwhelming performances, showing its strong generalization ability.}
Specifically, when GeCo is trained on FSCD-147, its performance drops significantly when evaluated on different datasets compared to when tested on FSCD-147 itself.
GeCo, like previous FSCD methods, relies on prototypes for both counting and detection.
We also suspect this lower generalization ability than ours originates from prototype matching, which is prone to overfitting to the training object semantics.
GeCo struggles when evaluated on datasets with different object semantics.
In contrast, \ours utilizes the structural information for matching instead of relying on semantic-intensive prototypes and generalizes more effectively on unseen datasets.

\begin{table}[t!]
\small
\centering
\scalebox{0.95}{
\begin{tabular}{lcccc}
\toprule
\multirow{2}{*}{Method}                     & \multicolumn{2}{c}{Seen}                                          & \multicolumn{2}{c}{Unseen}                                               \\  \cmidrule(lr){2-3} \cmidrule(lr){4-5}
                      & AP($\uparrow$) & AP50($\uparrow$) & AP($\uparrow$) & AP50($\uparrow$) \\ \midrule

FSDetView-PB~\cite{xiao2022few}            & 2.72  & 7.57  & 1.03  & 2.89\\
AttRPN-PB~\cite{fan2020few}               & 4.08  & 11.15 & 3.15  & 7.87 \\
C-DETR~\cite{cdeter}                 & 4.92  & 14.49 & 3.85  & 11.28 \\
DAVE~\cite{pelhan2024dave}                    & 6.75  & 22.51 & 4.12  & 14.16 \\
PseCo~\cite{pseco}                   & 22.37 & 42.56 & -     & -     \\
GeCo~\cite{geco}                    & -     & -     & 11.47 & 24.49 \\
\cellcolor{gray!20}\ours$_{\text{(ours)}}$                    & \cellcolor{gray!20}\textbf{27.49} & \cellcolor{gray!20}\textbf{48.48} & \cellcolor{gray!20}\textbf{22.71} & \cellcolor{gray!20}\textbf{39.68} \\
\bottomrule
\end{tabular}
}
\vspace{-2mm}
\caption{
Three-shot counting detection-based methods on the FSCD-LVIS seen and unseen split.
}
\label{tab:fscd-LVIS-results}
\vspace{-2.5mm}
\end{table}

\begin{table}[t!]
\small
    \centering
    \tabcolsep=0.16cm
    \scalebox{0.9}{
    \begin{tabular}{llccccc}
        \toprule
        & & & \multicolumn{2}{c}{AP} & \multicolumn{2}{c}{AP50} \\  \cmidrule(lr){4-5} \cmidrule(lr){6-7}
        Train  & Test & cross-eval & GeCo & TMR & GeCo & TMR \\ 
        \midrule
        \multirow{3}{*}{F-147}      & F-147                     &               &\textbf{43.42} & \cellcolor{gray!20}44.43          & \textbf{75.06}& \cellcolor{gray!20}73.83 \\
                                    & F-LVIS$_{\text{seen}}$    & \checkmark    & 13.96         & \cellcolor{gray!20}\textbf{21.25} & 25.87         & \cellcolor{gray!20}\textbf{37.18} \\
                                    & RPINE                     & \checkmark    & 19.47         & \cellcolor{gray!20}\textbf{26.21} & 38.69         & \cellcolor{gray!20}\textbf{52.01} \\ 
        \midrule
        \multirow{3}{*}{RPINE}      & F-147                     & \checkmark    & 36.99         & \cellcolor{gray!20}\textbf{41.39} & 60.38         & \cellcolor{gray!20}\textbf{69.19} \\
                                    & F-LVIS$_{\text{seen}}$    & \checkmark    & 10.01         & \cellcolor{gray!20}\textbf{20.92} & 17.44         & \cellcolor{gray!20}\textbf{37.87} \\
                                    & RPINE                     &               & 23.33         & \cellcolor{gray!20}\textbf{29.66} & 45.93         & \cellcolor{gray!20}\textbf{58.94}  \\
        \bottomrule
    \end{tabular}
    }
    \vspace{-2mm}
    \caption{Cross-dataset comparison of GeCo~\cite{geco} and \ours, where F-147, F-LVIS indicate FSCD-147 and FSCD-LVIS.
}
    \label{tab:cross-dataset}
    \vspace{-2.5mm}
\end{table}

\begin{table}[t!]
\small
\centering
\scalebox{0.95}{
\begin{tabular}{lcccc}
\toprule
\multirow{2}{*}{Method}                     & \multicolumn{2}{c}{One-shot}                                          & \multicolumn{2}{c}{Three-shot}                                               \\  \cmidrule(lr){2-3} \cmidrule(lr){4-5}
                      & AP($\uparrow$) & AP50($\uparrow$) & AP($\uparrow$) & AP50($\uparrow$) \\ \midrule
GeCo~\cite{geco}                    & 32.71 & 69.95     & 32.49 & 70.51 \\
\cellcolor{gray!20}\ours$_{\text{(ours)}}$                    & \cellcolor{gray!20}\textbf{36.01}   & \cellcolor{gray!20}\textbf{71.19} & \cellcolor{gray!20}\textbf{38.57}   & \cellcolor{gray!20}\textbf{72.61} \\
\bottomrule
\end{tabular}
}\vspace{-2mm}
\caption{Comparison without optional SAM decoder on FSCD147}
    \label{tab:no_sam_decoder}
    \vspace{-4mm}
\end{table}

\begin{table*}[ht]
\small
\centering
\vspace{-5mm}
\scalebox{0.92}{
\begin{tabular}{lccccccccc}
\toprule
\multirow{2}{*}{Method}& \multirow{2}{*}{SAM decoder} & \multicolumn{4}{c}{One-shot} & \multicolumn{4}{c}{Three-shot} \\  
\cmidrule(lr){3-6} \cmidrule(lr){7-10}
                     &    & MAE($\downarrow$) & RMSE($\downarrow$) & AP($\uparrow$) & AP50($\uparrow$) & MAE($\downarrow$) & RMSE($\downarrow$) & AP($\uparrow$) & AP50($\uparrow$) \\ 
\midrule
FSDetView-PB~\cite{xiao2022few}      &      & -     & -     & -     & -     & 37.54 & 147.07& 13.41 & 32.99 \\
AttRPN-PB~\cite{fan2020few}        &      & -     & -     & -     & -     & 32.42 & 141.55& 20.97 & 37.19 \\
C-DETR$\dagger$~\cite{cdeter}            &     & 16.99 & 125.22 & 19.14 & 47.63 & 16.79 & 123.56& 22.66 & 50.57 \\
SAM-C$\dagger$ ~\cite{ma2023samcountanythingempirical}            & \checkmark     & 33.17 & 141.77 & 35.09 & 56.02 & 27.97 & 131.24& 27.99 & 49.17 \\
PseCo~\cite{pseco}             & \checkmark     & 14.86 & 118.64 & 41.63 & 70.87 & 13.05 & 112.86& 42.98 & 73.33 \\
DAVE~\cite{pelhan2024dave}              &   & \underline{11.54}  & 86.62 & 19.46 & 55.27 & \underline{10.45} & 74.51 & 26.81 & 62.82 \\
GeCo~\cite{geco}              &  \checkmark    & \textbf{8.10}  & \underline{60.16} & \underline{43.11} & \textbf{74.31} & \textbf{7.91}  & \underline{54.28} & \underline{43.42} & \textbf{75.06} \\
\cellcolor{gray!20}\ours$_{\text{(ours)}}$ &     \cellcolor{gray!20}\checkmark  & \cellcolor{gray!20}11.63     & \cellcolor{gray!20}\textbf{57.46}     & \cellcolor{gray!20}\textbf{43.15}   & \cellcolor{gray!20}\underline{71.55}   & \cellcolor{gray!20}13.78  & \cellcolor{gray!20}\textbf{51.87}& \cellcolor{gray!20}\textbf{44.43}   & \cellcolor{gray!20}\underline{73.83}   \\
\bottomrule
\end{tabular}
}
\vspace{-2mm}
\caption{
Comparison with few-shot detection and counting methods on FSCD-147. The one-shot performance of the models with $\dagger$ are evaluated using the official code.
}
\label{tab:fsc147-results}
\vspace{-3mm}
\end{table*}

\begin{figure*}[t!]
	\centering
	\small
    \includegraphics[width=0.99\linewidth]{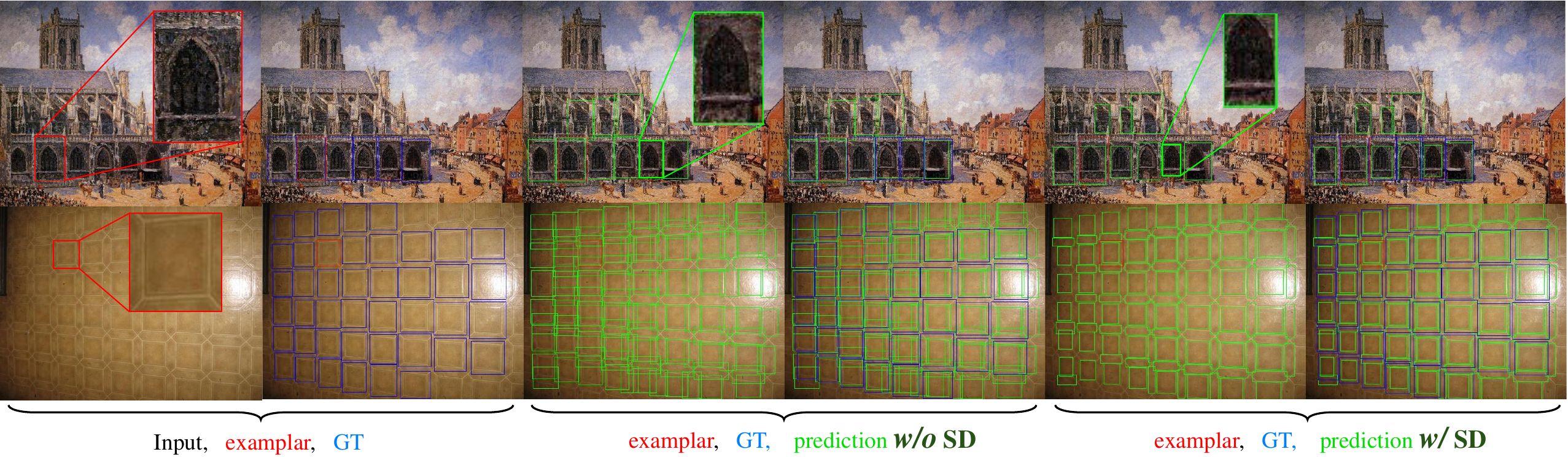}
    \vspace{-2mm}
 \caption{\ours without \textit{vs.}~with SAM decoder~\cite{sam} (\textit{w/o} SD \textit{vs.}~\textit{w/} SD) for box refinement. SAM decoder tends to align box predictions with the nearest edges. The edge-sensitive prediction is harmful for non-objects, objects that have loose boundaries, or patterns with many confusing edges. The exemplar in the first row loosely contains an arched window, but the SAM decoder snaps the box tightly its edges.
 }
\label{fig:sam_failures}
\vspace{-5mm}
\end{figure*}

\begin{table}[t!]
\small
    \centering
    \scalebox{0.9}{
    \begin{tabular}{llcccc}
        \toprule
        & \multirow{2}{*}{feature $\mF_{\text{P}}$} & \multicolumn{2}{c}{RPINE} & \multicolumn{2}{c}{FSCD-147} \\  \cmidrule(lr){3-4} \cmidrule(lr){5-6}
        &  & AP & AP50 & AP & AP50 \\ 
        \midrule
        (a) & $\mF$                             & 11.44 & 24.02 & 20.95 & 55.59 \\
        (b) & $\mF_{\text{TM}}$                 & 32.55 & 57.43 & 31.96 & 63.03 \\\midrule
        (c) & $\mF \oplus \mF_{\text{TM-cos}}$  & 29.74 & 57.85 & 24.73 & 60.29 \\ 
        (d) & $\mF \oplus \mF_{\text{PM}}$      & 20.94 & 47.96 & 28.91 & 66.93 \\\midrule
        (e) & $\mF \oplus \mF_{\text{TM}}$      & \textbf{33.59} & \textbf{64.05} & \textbf{36.01} & \textbf{71.19} \\ 
        \bottomrule
    \end{tabular}
    }
    \vspace{-2mm}
    \caption{Effect of template matching features (Sec.~\ref{sec:tm}).
    $\mF_{\text{PM}}$ is the correlation with average-pooled prototype.
    $\mF_{\text{TM-cos}}$ is the cosine similarity of the template and image feature.
}
    \label{tab:ablation_tm}
    \vspace{-6mm}
\end{table}

\smallbreakparagraph{SAM decoder is biased to edges.}
Table~\ref{tab:RPINE-1shot-results} shows the negative impact of the optional box refinement using the SAM decoder~\cite{sam} on RPINE.
Figure~\ref{fig:sam_failures} visualizes two representative examples when the SAM decoder~(SD) degrades performance.
We observe that SD tends to align box predictions with the nearest edge, which is expected given that SAM is a segmentation model.
The edge-sensitive prediction is particularly harmful for non-typical objects, such as objects that are loosely bounded by edges or patterns with many confusing edges.
The SD refinement is seemingly good at edge detection for typical object exemplars with clear boundaries, and this is why the existing FSCD methods~\cite{pseco, geco, ma2023samcountanythingempirical} benefit from adopting SD for box refinement.
However, arbitrary patterns are not necessarily bounded by clear edges.
Hence, edge-driven refinement may be even harmful as verified on \ourdataset.
We emphasize that SD is an additional box post-processor, and we optionally add SD to \ours to compare with the SD-based state of the arts.
Table~\ref{tab:no_sam_decoder} compares models by taking out SD, where \ours shows powerful performance.

\smallbreakparagraph{Effectiveness of template matching features.}
We verify the effectiveness of 2D template matching, which preserves the spatial structure of exemplar bounding boxes for correlation.
We experiment with different input feature $\mF_{\text{P}}$ to the box regression module.
Table~\ref{tab:ablation_tm} compares our final model (e) and its variants.
The lower bound model (a) observes zero information on the exemplar. 
Model (b) shows that correlating with 2D templates is greatly beneficial.
The final model (e) concatenates the template-matching and the image feature, providing the correlation and appearance information for pattern matching regression.
The models (c, d) replace the variants of the template matching feature $\mF_{\text{TM}}$ with something else.
Model (c) replaces the channel-wise correlation of (e) to cosine similarity, \ie, $\mF_{\text{TM-cos}}$, showing that retaining the channel dimension helps.
Model (d) first performs average-pooling of the template and uses the prototype matching feature, $\mF_{\text{PM}}$, for correlation. Note that this average-pooled feature is often used in previous FSC methods~\cite{you2022iterative, pelhan2024dave, geco}.
The result shows that preserving the spatial structure of exemplar bounding boxes is more effective.
The representative failure cases are shown in Fig.~\ref{fig:pm_failures} in supp. material.
The pooled prototype feature loses the geometric layout of the template and struggle to detect patterns where geometric clues are crucial.
Comparing (d) and (e) verifies our hypothesis that prototype-based matching relies on the object priors;
Model (d) drops more significantly on \ourdataset, which is less object-centric than FSCD-147.

\begin{table}[t!]
\small
    \centering
    \scalebox{0.88}{
    \tabcolsep=0.15cm
    \begin{tabular}{lcccccc}
        \toprule
        & \multicolumn{2}{c}{box regression} & \multicolumn{2}{c}{RPINE} & \multicolumn{2}{c}{FSCD-147} \\  \cmidrule(lr){2-3} \cmidrule(lr){4-5} \cmidrule(lr){6-7}
        & shift & scale & AP & AP50 & AP & AP50 \\ 
        \midrule
        (a) & $(0,0)$                      & $(0,0)$                               & 26.20 & 55.20 & 22.81 & 57.30 \\
        (b) & $(\Delta x, \Delta y)$       & $(e^{\alpha_w}, e^{\alpha_h})$        & 23.88 & 51.25 & 17.01 & 55.14 \\
        (c) & $(\Delta x, \Delta y)$       & $(e^{\alpha_w}s_w, e^{\alpha_h}s_h)$  & 31.29 & 59.52 & 35.08 & 70.23 \\
        \midrule
        (d) & $(s_w\Delta x, s_h\Delta y)$ & $(e^{\alpha_w}s_w, e^{\alpha_h}s_h)$  & \textbf{33.59} & \textbf{64.05} & \textbf{36.01} & \textbf{71.19} \\
        \bottomrule
    \end{tabular}
    }\vspace{-2mm}
    \caption{Effect of box regression methods (Sec.~\ref{sec:tr})
}
    \label{tab:ablation_tr}
\vspace{-5mm}
\end{table}

\smallbreakparagraph{Effectiveness of support-conditioned regression.}
Table~\ref{tab:ablation_tr} compares different variants for box regression and validates the effectiveness of our support-conditioned box regression method.
As shown in (b), regressing bounding boxes without considering the support exemplar's size leads to poor localization.
Notably, it even performs worse than (a), which directly uses the exemplar box without performing any regression.
However, as shown in (c), incorporating the support exemplar's size when predicting the width and height leads to a significant performance gain.
This improvement is further enhanced by using the exemplar's size for shifting, as demonstrated in (d).
These results demonstrate the effectiveness of support-conditioned regression, guided by the exemplar's size, in achieving accurate localization.

\begin{table}[t!]
\small
    \centering
    \scalebox{0.98}{%
    \begin{tabular}{lccc}
    \toprule
     Method &  Trainable params  & Total params  & FLOPS \\
     \midrule 
     PseCo~\cite{pseco} & 56.99M & 0.70B & 5.08T \\
     GeCo~\cite{geco} &  7.98M & 0.65B & 4.72T \\
     \cellcolor{gray!20}\ours$_{\text{(ours)}}$ & \cellcolor{gray!20}19.01M &  \cellcolor{gray!20}0.66B & \cellcolor{gray!20}3.04T \\
     \bottomrule
    \end{tabular}
    }\vspace{-2mm}
\caption{Comparison on the computational complexity}
    \label{tab:flops}
    \vspace{-4mm}
\end{table}

\smallbreakparagraph{Comparison on computational complexity.}
Table~\ref{tab:flops} compares the complexity of \ours with state-of-the-art FSCD methods, demonstrating its efficiency and effectiveness.
Compared to PseCo~\cite{pseco}, which introduces a large number of trainable parameters and FLOPs, \ours introduces only about 19M trainable parameters, which is detailed in Tab.~\ref{tab:architecture_details}.
Although \ours has more trainable parameters than GeCo~\cite{geco}, its total parameter count remains comparable. 
Thanks to the simple architecture, \ours is significantly efficient in terms of FLOPs (3.04T), which is notably lower than PseCo (5.08T) and GeCo (4.72T).
This reduction in computational cost not only enhances training efficiency but also results in faster inference. In contrast, the high FLOPs of PseCo and GeCo contribute to longer inference, making them less practical for real-time applications.

\begin{figure}[t!]
	\centering
	\small    \includegraphics[width=0.99\linewidth]{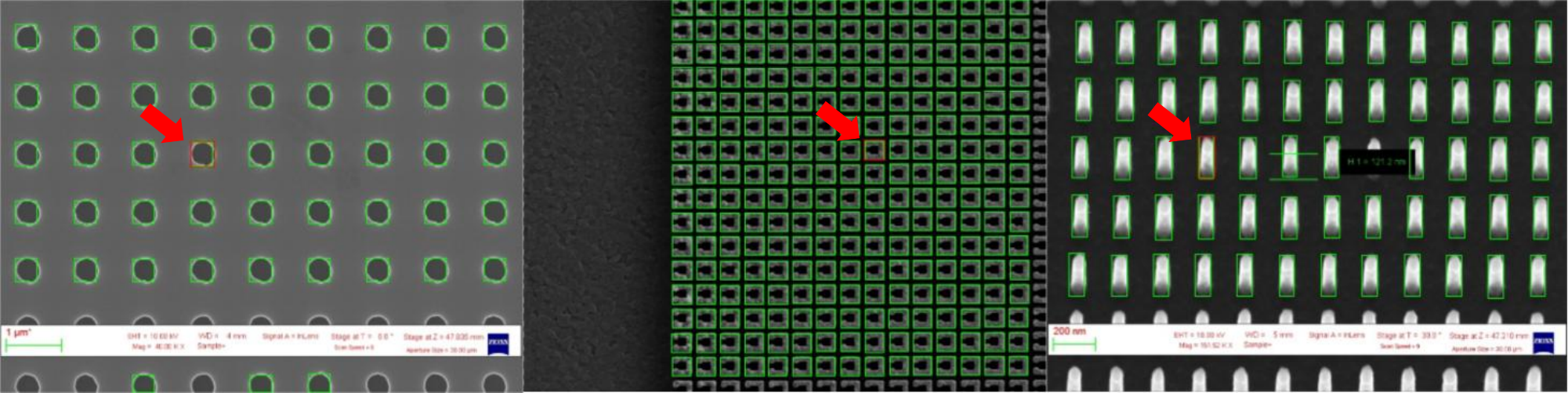}
        \vspace{-2mm}
 \caption{Application example of \ours on the SEM dataset~\cite{aversa2018first}.}
\label{fig:qual_SEM2}
\vspace{-5mm}
\end{figure}

\smallbreakparagraph{Real-world application.}
We evaluate \ours, trained on RPINE, on scanning electron microscope (SEM) images~\cite{aversa2018first}, which are widely used in microprocessor inspection.
Despite the domain shift, our method performs effectively and demonstrates its potential to generalize to real-world, non-object pattern scenarios (Fig.~\ref{fig:qual_SEM2}).


\section{Conclusion and discussion}
We have proposed a simple template-matching based method for few-shot pattern detection.
We also introduce a new dataset with bounding box annotations that covers various patterns around the world across non-object patterns to typical objects from various image domains from nature and human-made products.

Our experiments show that integrating the widely used SAM decoder improves performance by 7\% AP on FSCD-147 via object-level edge priors, but causes a drop on RPINE, which contains more general, non-object patterns. 
This highlights that strong object priors can benefit object-like patterns but may hinder general pattern detection.

Future work could explore combining the powerful pre-trained knowledge of foundation models with detection modules less reliant on object-level edge priors.
Another direction is to design lightweight pattern-specific architectures with minimal object priors to better capture fine-grained repetitive structures and improve generalization.

\vspace{-3mm}
\paragraph{Acknowledgments.}
This work was supported by Samsung Electronics Co., Ltd (IO240508-09825-01) and IITP grants (RS-2022-II220959: Few-Shot Learning of Causal Inference in Vision \& Language (50\%), RS-2022-II220113: Developing a Sustainable Collaborative Multi-modal Lifelong Learning Framework (45\%), RS-2019-II191906: AI Graduate School Program at POSTECH (5\%)) funded by Ministry of Science and ICT, Korea.
We also appreciate Manjin Kim's advice on constructing RPINE.

{
    \small
    \bibliographystyle{ieeenat_fullname}
    \bibliography{main}
}
\clearpage
\setcounter{page}{1}
\maketitlesupplementary

In this supplementary material, we provide additional experimental results and analysis to support our method, including qualitative results.
\section{Additional details}
\subsection{Detailed model architecture}
We provide the details of the model architecture in Tab.~\ref{tab:architecture_details}.
We design our model architecture to be as simple as possible, and there are only 6 learnable layers in total.
\begin{table}[h!]
\small
    \centering
    \tabcolsep=0.13cm
    \scalebox{0.92}{%
    \begin{tabular}{llr}
    \toprule
     module & structure & \# params. \\ \midrule
     backbone projection & linear(in=256, out=512) & 0.13M \\
     $\mF_{\text{TM}}$ scaler & nn.Parameter & 1 \\ \midrule
     \multirow{3}{*}{box regressor} & conv(k=(3, 3), in=1024, out=1024)  & 9.44M \\
      & LeakyReLU  & 0 \\
      & linear(in=1024, out=4) & 4096 \\\midrule
     \multirow{3}{*}{presence classifier} & conv(k=(3, 3), in=1024, out=1024)  & 9.44M \\
     & LeakyReLU  & 0 \\
     & linear(in=1024, out=1) & 1024 \\
     \bottomrule
    \end{tabular}
    }
\caption{The total learnable layers in \ours. We exclude the feature backbone parameters that are frozen. The ``k'' in the table denotes the 2D convolution kernel size.}
    \label{tab:architecture_details}
    \vspace{-3mm}
\end{table}

\subsection{Template extraction details}
\label{sec:supp_template_extraction}

\begin{figure}[h!]
\vspace{-4mm}
\begin{center}
   \includegraphics[width=0.99\linewidth]{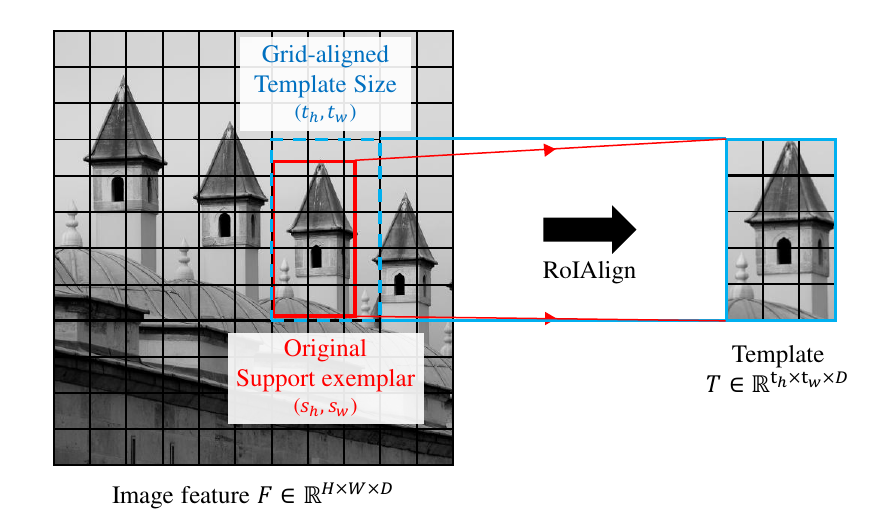}     
\end{center}
\vspace{-6mm}
\caption{
Using RoIAlign, the template $\mT$ (blue box) is extracted by adaptively determining its size ($t_h, t_w$) (blue dashed box) to fully cover the support exemplar's region ($s_h, s_w$) (red box) on $\mF$, which preserves spatial alignment between $\mF$ and $\mT$.
} 
\label{fig:TE}
\vspace{-2mm}
\end{figure}

Prior methods typically use Global Average Pooling or RoIAlign to produce fixed-size prototypes.
However, these approaches can lead to spatial misalignment between the feature map $\mF$ and the template $\mT$, which degrades template matching performance.
As shown in Fig.~\ref{fig:TE}, we address this issue by adaptively determining the template size ($t_h, t_w$) based on the support exemplar's region ($s_h, s_w$) on $\mF$ (red box), rounding it up to the smallest grid-aligned region that fully contains the support exemplar's area (blue dashed box).
Using this size, we apply RoIAlign to extract a spatially aligned template (blue box), enabling more precise and consistent template matching.

\subsection{Definition of the extended center point set}
\label{sec:supp_xp_details}
To avoid supervising the presence prediction with only a single pixel at the ground-truth center, we define $\mathcal{X}_{\text{P}}$ as a set of extended center point within a margin $\delta$ around each ground-truth center point $(x_\text{c}, y_\text{c})$.

\begin{figure}[t!]
	\centering
	\small
    \centering
    \includegraphics[width=0.99\linewidth]{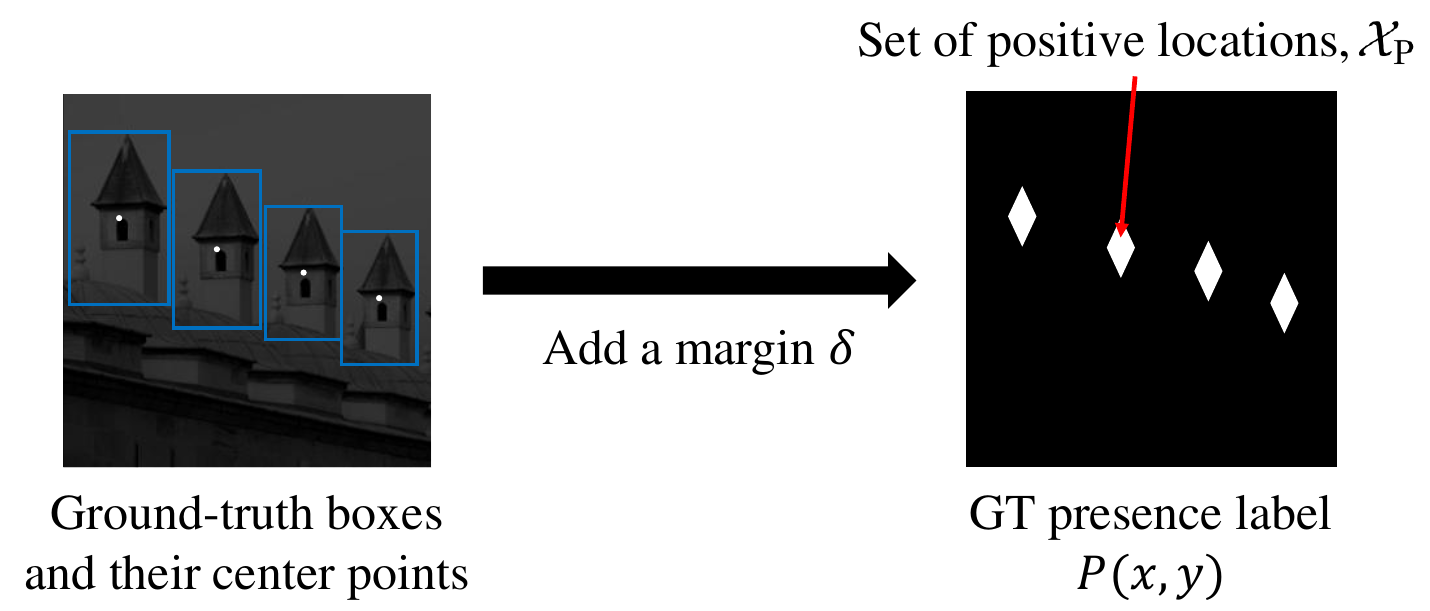}
    \vspace{-2mm}
 \caption{Definition of $\mathcal{X}_{\text{P}}$ using GT centers and margin $\delta$.
 }    \label{fig:margin}

\vspace{-4mm}
\end{figure}

A location $(x, y)$ is considered positive if it falls within this margin region around the center of any ground-truth bounding box.
As illustrated in Fig.~\ref{fig:margin}, $\mathcal{X}_{\text{P}}$ is defined as follows:
\small
\begin{equation}
    \mathcal{X}_{\text{P}} = \left\{ (x,y) \;\middle|\; \forall (x_\text{c}, y_\text{c}, w, h) \in \mathcal{B},\ 
    \frac{|x_\text{c} - x|}{w} + \frac{|y_\text{c} - y|}{h} \leq \delta \right\}.
\end{equation}
\normalsize
Here, $\mathcal{B}$ denotes the set of ground-truth boxes; $(x_\text{c}, y_\text{c})$ represents the center coordinates, and $(w, h)$ the width and height of each ground-truth box.
The resulting shape of $\mathcal{X}_{\text{P}}$ forms a rhombus centered at each ground-truth location.
We fix $\delta = 0.33$ in all experiments.

\subsection{Dataset details}
\textbf{FSCD-147} extends FSC-147~\cite{ranjan2021learning} dataset, which includes only dot annotations for objects, to include bounding box annotations.
It covers 147 object categories with additional bounding box annotation.
\textbf{FSCD-LVIS} includes more complex scenes with multiple object classes, each containing multiple instances, compared to FSCD-147, where each image has a relatively simple scene.
Regardless, FSCD-LVIS still uses a single pattern per image unlike \ourdataset that is annotated with multiple existing pattern classes.

\section{Additional experimental details}

\begin{figure}[t!]
	\centering
        \vspace{-2mm}
	\small    \includegraphics[width=0.99\linewidth]{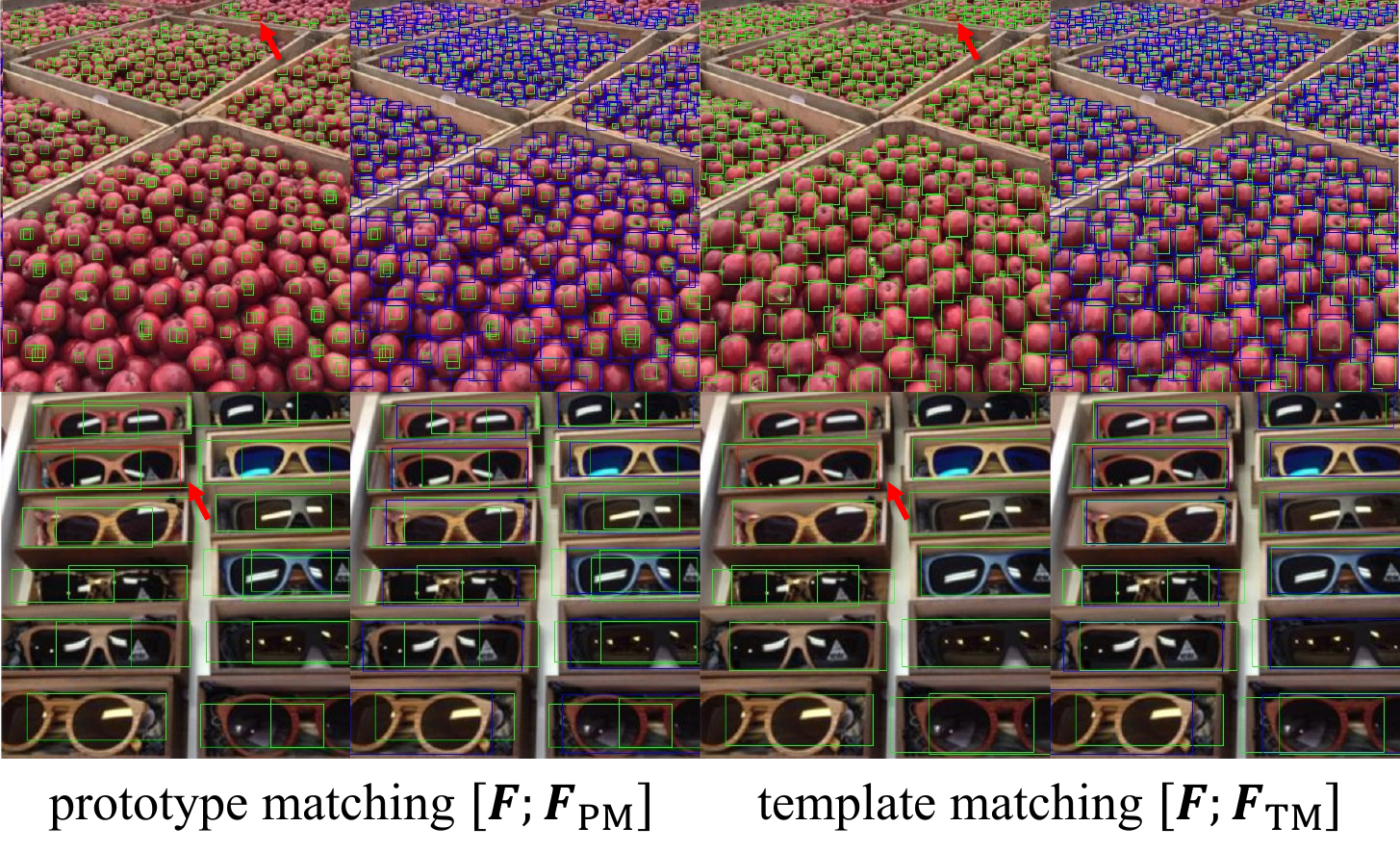}
        \vspace{-2mm}
 \caption{Noticeable failure cases of prototype matching. Prototypes collapse the geometric layout of the exemplars, being especially vulnerable for localizing instances among dense repetition or patterns including sub-patterns.
 }
\label{fig:pm_failures}
\vspace{-2mm}
\end{figure}

\begin{figure}[t!]
\vspace{-4mm}
\begin{center}
   \includegraphics[width=0.99\linewidth]{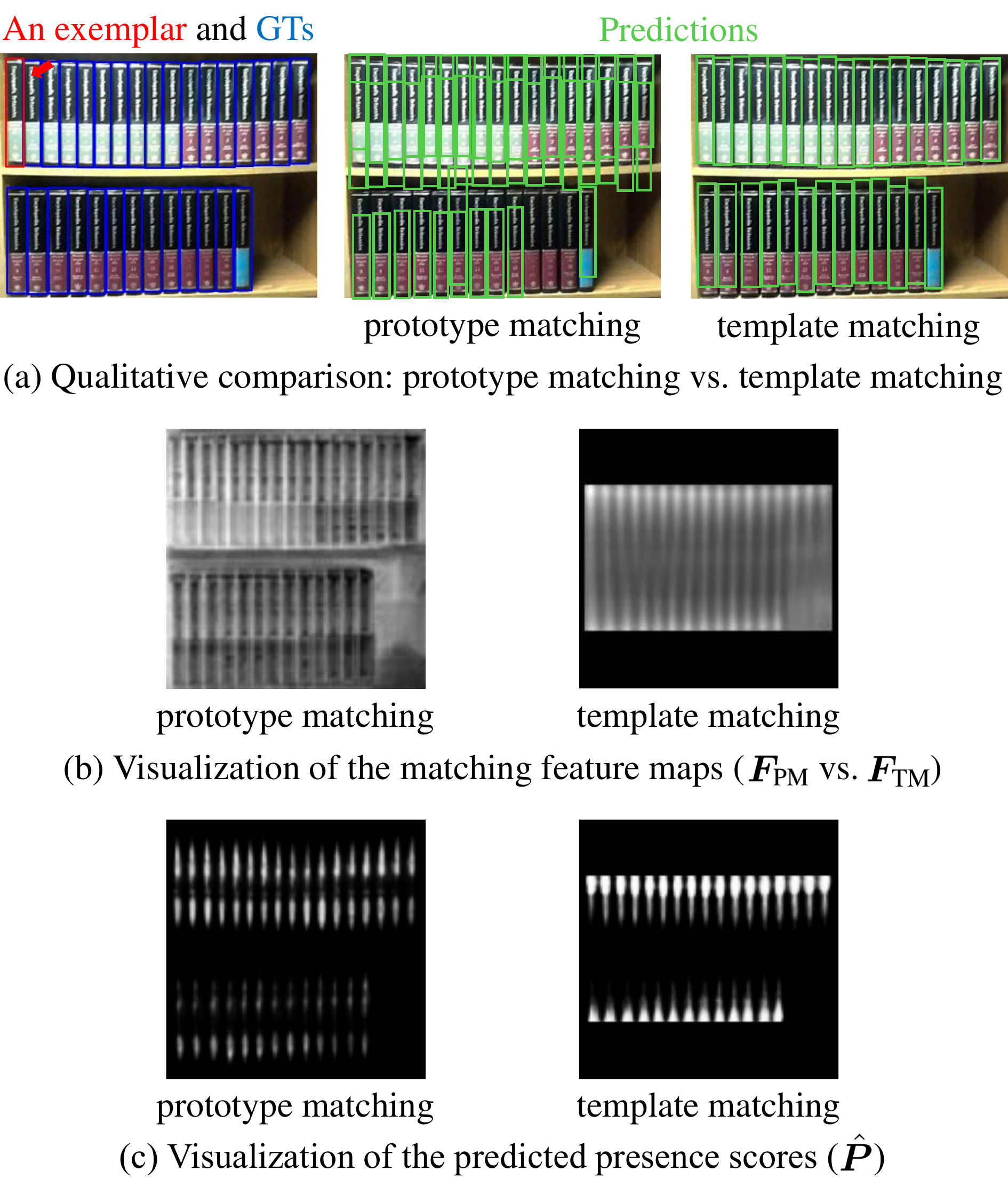}     
\end{center}
\vspace{-6mm}
\caption{
Comparison between prototype matching and template matching.
} 
\label{fig:prototype_analysis}
\vspace{-4mm}
\end{figure}

\paragraph{Qualitative analysis of prototype matching failures.}
Fig.~\ref{fig:pm_failures} shows failure cases of the prototype matching method compared to our template matching approach.
Fig.~\ref{fig:prototype_analysis} provides a detailed comparison in terms of box regression (a), matching feature maps (b), and presence scores (c).
In both Fig.~\ref{fig:pm_failures} and Fig.~\ref{fig:prototype_analysis} (a), 
prototype-matching models often produces inaccurate bounding boxes that fail to tightly enclose the target pattern.

Fig.~\ref{fig:prototype_analysis} (b) and (c) further illustrate the differences in feature maps and predicted presence scores, respectively.
Prototype feature maps ($\mF_\text{PM}$) highlight regions with similar semantics (e.g., edges or colored bends of a book) while ignoring the exemplar’s spatial structure, making it difficult to localize the center of the target pattern.
In contrast, template matching feature maps ($\mF_\text{TM}$) preserve spatial structure and clearly emphasize the central region of the target pattern, enabling more precise localization.
Consistent with this, the predicted presence score maps in Fig.~\ref{fig:prototype_analysis} (c) show that prototype-based scores often activate spatially misaligned but semantically related regions, while template-based scores focus accurately on the true target center.
These observations highlight the limitations of prototype matching in capturing spatial structure and demonstrate the effectiveness of template matching for precise localization.

\begin{figure}[t!]
\begin{center}
   \includegraphics[width=0.99\linewidth]{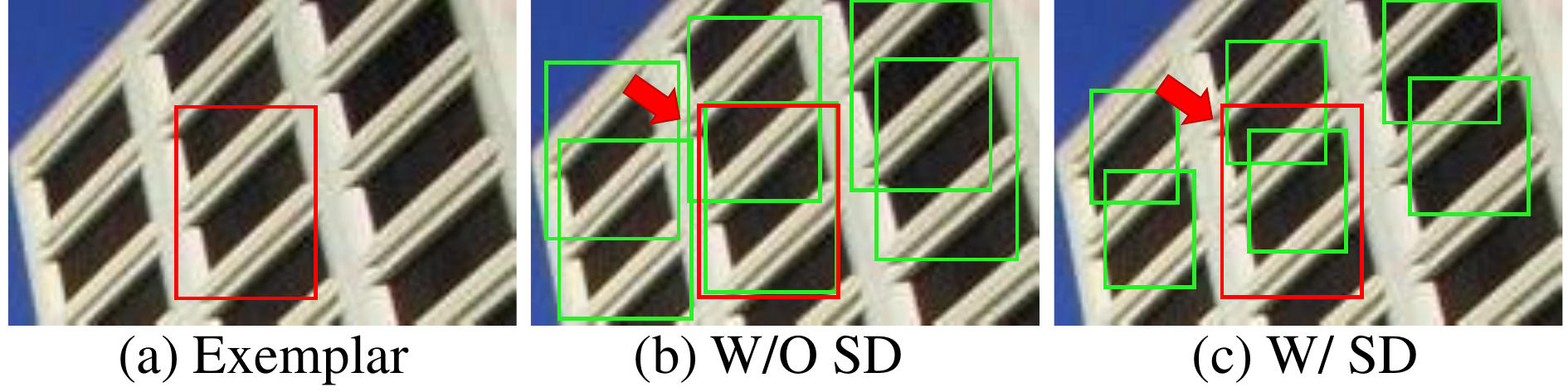}     
\end{center}
\vspace{-6mm}
\caption{
Additional failure case of SAM decoder.
} 
\label{fig:SD_more}
\vspace{-4mm}
\end{figure}

\begin{figure*}[t!]
\centering
\vspace{-4mm}
\small
\scalebox{0.85}{
    \includegraphics[width=0.95\linewidth]{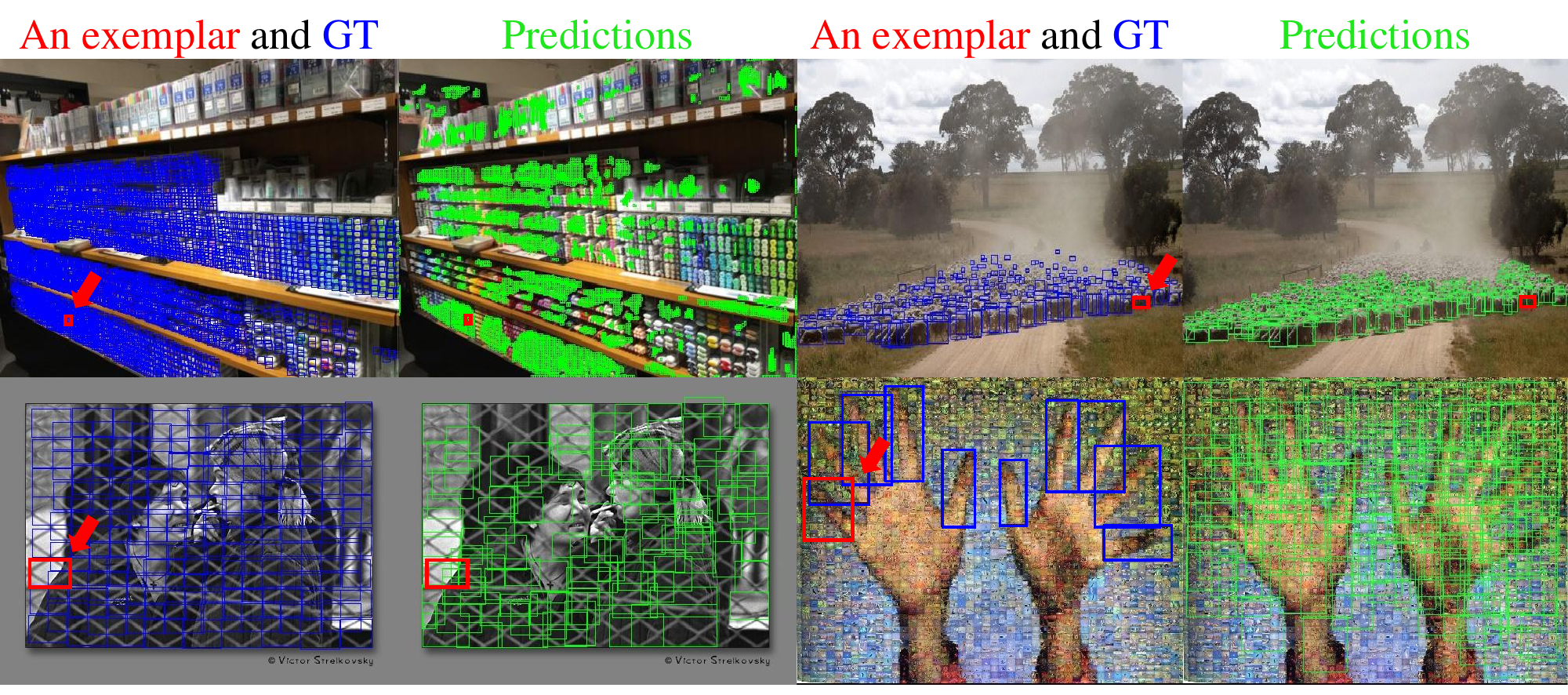}
}
\vspace{-2mm}
\caption{Qualitative analysis of failure cases in TMR.} 
\label{fig:failures}
\vspace{-6mm}
\end{figure*}

\vspace{-2mm}
\paragraph{Failure cases of TMR.}
We analyze the failure cases of TMR, as shown in Fig.~\ref{fig:failures}.
In the first row of Fig.~\ref{fig:failures}, TMR often fails to detect highly crowded patterns when the support exemplar is extremely small.
In the second row, the model struggles with highly textured patterns that exhibit large variations in texture appearance.
These examples suggest potential directions for improvement, such as incorporating multi-scale representations with much higher resolution features and designing pattern-specific backbones to reduce noise caused by texture variation.

\vspace{-2mm}
\paragraph{Additional analysis of SAM decoder's (SD) failures.}
To support our main analysis in Sec.\ref{sec:ablation}, we provide additional qualitative example highlighting the limitations of the SAM decoder (SD) in handling non-object patterns. 
As shown in Fig.\ref{fig:SD_more}, although the exemplar includes both the window and decoration around the window sills, the refined prediction closely aligns with the black window frames, missing the broader structure present in the exemplar.
This edge-sensitive behavior of SD is consistent with the findings in the Sec.{\color{iccvblue}7.2} of the SAM paper~\cite{sam}, which reports that SD produces high-recall edge maps even without explicit edge supervision.

\begin{figure*}[h!]
	\centering
        \vspace{-2mm}
	\small    \includegraphics[width=0.8\linewidth]{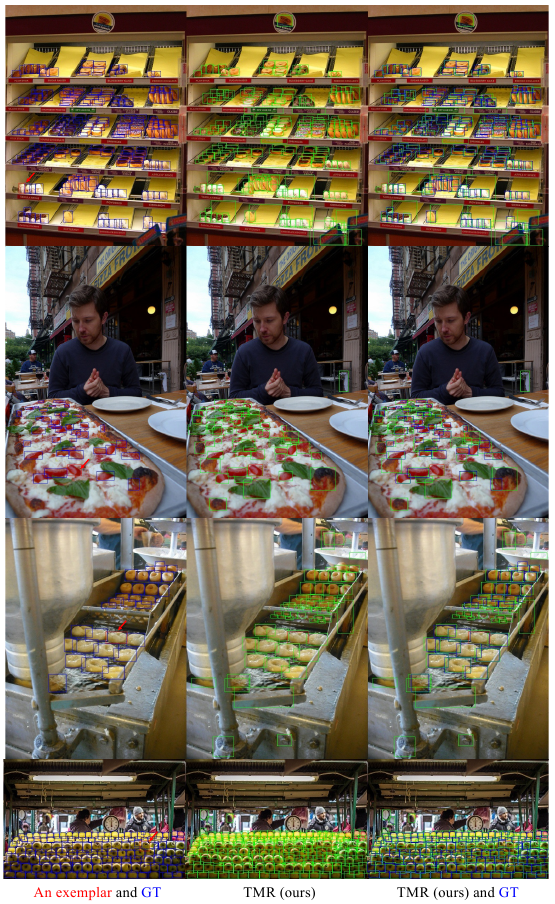}
        \vspace{-2mm}
 \caption{Additional qualitative results on the FSCD-LVIS dataset.}
\label{fig:sup_qual_lvis}
\vspace{-4mm}
\end{figure*}

\begin{figure*}[h!]
	\centering
        \vspace{-2mm}
	\small    \includegraphics[width=0.99\linewidth]{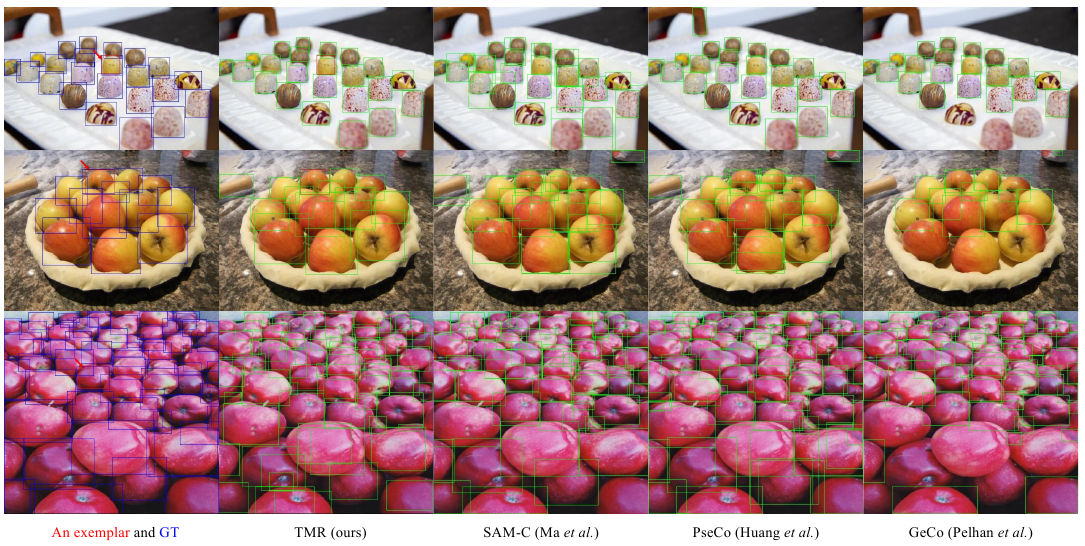}
        \vspace{-2mm}
 \caption{Additional qualitative results on the FSCD-147 dataset.
 }
\label{fig:sup_qual_fcds}
\vspace{-4mm}
\end{figure*}

\begin{figure*}[h!]
	\centering
        \vspace{-2mm}
	\small    \includegraphics[width=0.99\linewidth]{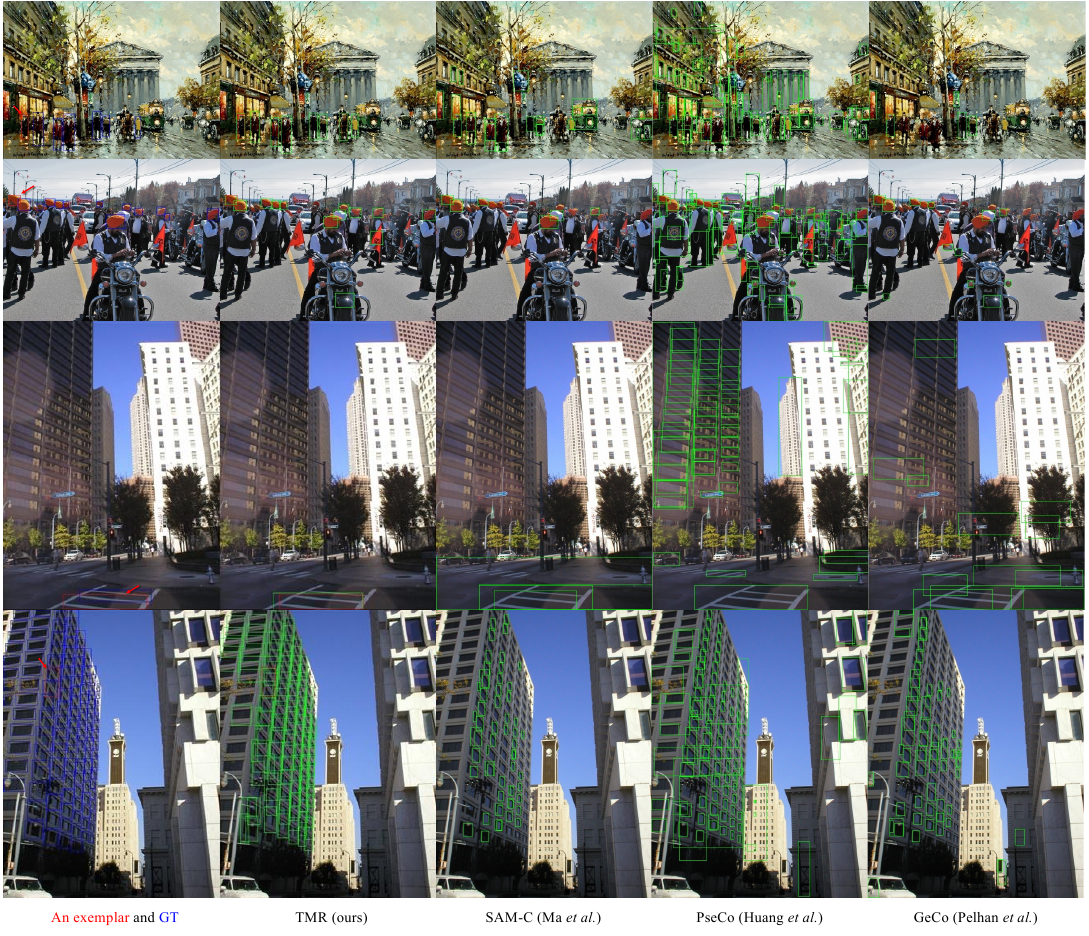}
        \vspace{-2mm}
 \caption{Additional qualitative results on the RPINE dataset.}
\label{fig:sup_qual_rpine}
\vspace{-4mm}
\end{figure*}

\vspace{-2mm}
\paragraph{Qualitative results on FSCD-147, FSCD-LVIS and RPINE.}
We provide additional qualitative results on the FSCD-147, FSCD-LVIS and RPINE dataset in Figs.~\ref{fig:sup_qual_fcds}, Figs.~\ref{fig:sup_qual_lvis} and~\ref{fig:sup_qual_rpine} to show the model's effectiveness.
As shown in Fig.~\ref{fig:sup_qual_fcds}, our method effectively detects the given exemplar. For instance, in the first row, \ours successfully identifies all instances of the given exemplar, whereas other state-of-the-art models either fail to detect some instances or produce false positives.
Furthermore, \ours successfully detects non-object patterns, as shown in the penultimate row of Fig.~\ref{fig:sup_qual_rpine}.

\paragraph{Multi-scale extension.}
\label{sec:multi-scale}
\ours can be naturally extended to multi-scale prediction by incorporating a multi-scale architecture, such as ViTDet~\cite{li2022exploringplainvisiontransformer}.
Following a similar approach to the few-shot extension in Sec.~\ref{sec:inference}, we first extract feature maps at multiple scales and independently apply the prediction process to each scale.
The resulting predictions are then aggregated and filtered using Non-Maximum Suppression (NMS) to remove duplicates across scales.
As shown in Tab.~\ref{tab:RPINE-multiscale}, leveraging multi-scale features yields further performance improvements, demonstrating the benefit of scale-aware detection in capturing pattern instances of varying sizes.
For a fair comparison with prior state-of-the-art methods~\cite{geco, pseco}, we use the single-scale setting in all experiments, except for the multi-scale experiment reported in Tab.~\ref{tab:RPINE-multiscale}.

\begin{table}[t!]
\centering
\small
\setlength{\tabcolsep}{5pt}
\scalebox{0.9}{
\tabcolsep=0.2cm
\begin{tabular}{lcccccc}
\toprule
Type& \multicolumn{3}{c}{Scales} & AP & AP50 & AP75 \\  \cmidrule(lr){2-4}
& $32^2$ & $64^2$ & $128^2$ & & & \\
\midrule
\multirow{2}{*}{Single-Scale}& & \checkmark &            & 27.49 & 56.15  & 23.25 \\
& \cellcolor{gray!20}& \cellcolor{gray!20}& \cellcolor{gray!20}\checkmark    & \cellcolor{gray!20}33.59 & \cellcolor{gray!20}64.05   & \cellcolor{gray!20}30.52 \\
\midrule
\multirow{3}{*}{Multi-Scales}& \checkmark & & \checkmark       & 34.03 & 64.86   & 31.66 \\
& & \checkmark & \checkmark      & 34.78 & 66.71   & 31.51 \\
& \checkmark & \checkmark & \checkmark  & \textbf{35.41} & \textbf{66.88}   & \textbf{32.52} \\
\bottomrule
\end{tabular}
}
\vspace{-2mm}
\caption{Multi-scales experiment of \ours on RPINE. The gray-shaded row indicates the default single-scale configuration described in Sec.~\ref{sec:method}, where only a single feature scale is used.
For the single-scale setting, upscaling the feature map from $64\times64$ to $128\times128$ improves performance, as the resulting higher-resolution correlation map enables denser predictions.
}
\vspace{-4mm}
\label{tab:RPINE-multiscale}
\end{table}

\begin{figure}[t!]
	\centering
	\small    \includegraphics[width=0.99\linewidth]{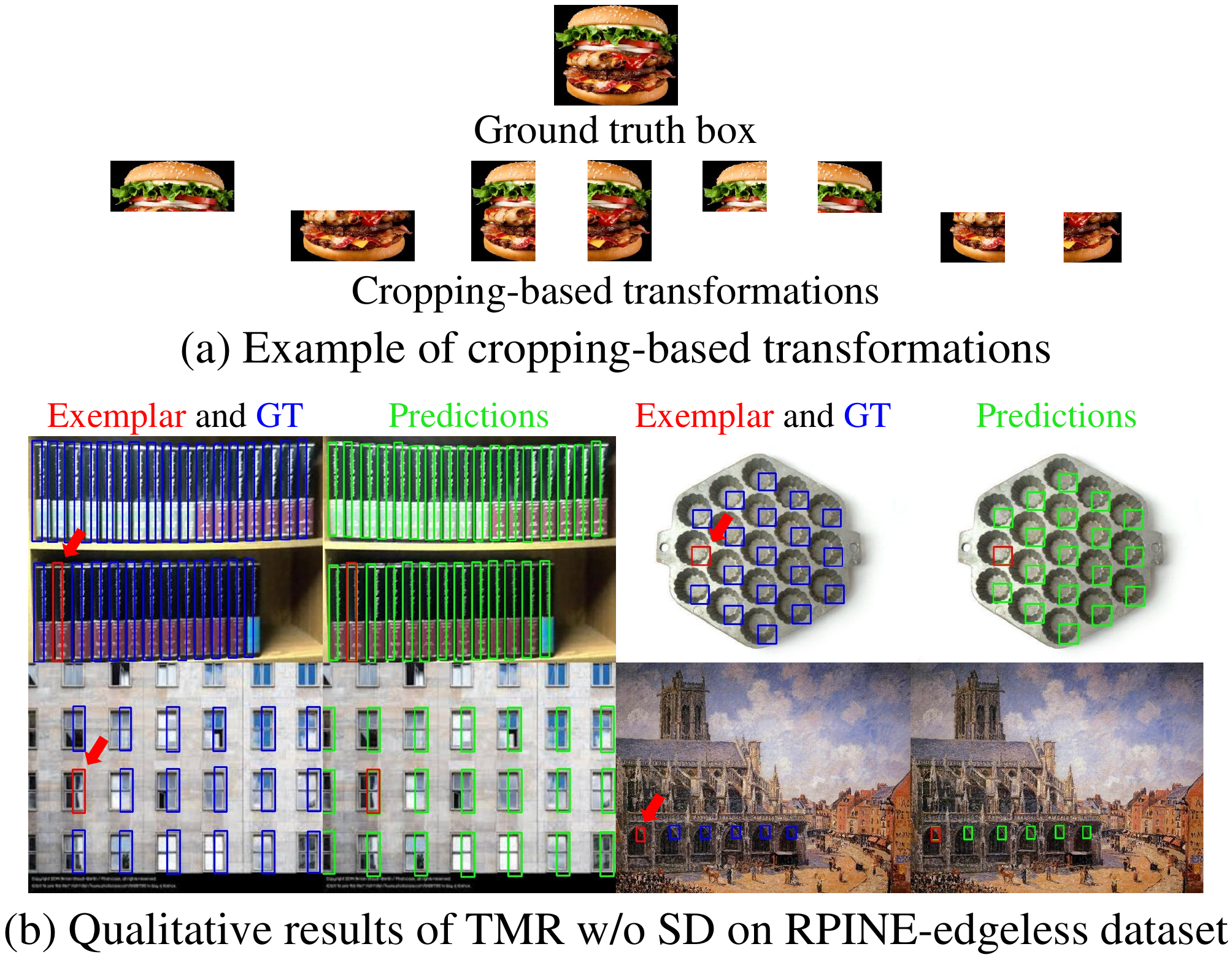}
        \vspace{-3mm}
 \caption{RPINE-edgeless examples.}
\label{fig:sup_RPINE_edgeless}
\vspace{-2mm}
\end{figure}

\vspace{-2mm}
\paragraph{RPINE-edgeless.}
\label{sec:RPINE-edgeless}
To evaluate the model under the minimal assumptions of the object-level edge prior, we augment the RPINE dataset via bounding box transformation.  
Given a ground-truth bounding box, we apply eight types of cropping-based transformations: left half, right half, top half, bottom half, top-left corner, top-right corner, bottom-left corner, and bottom-right corner, as illustrated in Fig.~\ref{fig:sup_RPINE_edgeless} (a).  
Based on this, we construct the RPINE-edgeless dataset (Fig.~\ref{fig:sup_RPINE_edgeless} (b)), which contains 13,772 training samples and 1,402 validation samples.  
As shown in Tab.~\ref{tab:RPINE-edgeless}, we compare \ours with GeCo, and \ours demonstrates strong performance.  
However, we also observe that when using SD, the performance drops significantly, which aligns with our claim in Sec.~\ref{sec:ablation}.

\begin{table}[t!]
\small
\centering
\scalebox{0.9}{
\begin{tabular}{lcccc}
\toprule
Method   & SD & AP($\uparrow$) & AP50($\uparrow$) & AP75($\uparrow$) \\ \midrule
GeCo~\cite{geco}              & \checkmark & 12.53 & 30.96 & 8.55 \\
\cellcolor{gray!20}\ours$_{\text{(ours)}}$     & \cellcolor{gray!20} & \cellcolor{gray!20}\textbf{33.25} & \cellcolor{gray!20}\textbf{67.63} & 
\cellcolor{gray!20}\textbf{28.22} \\
\cellcolor{gray!20}\ours$_{\text{(ours)}}$      & \cellcolor{gray!20}\checkmark & \cellcolor{gray!20}\textbf{17.99} & \cellcolor{gray!20}\textbf{46.31} & \cellcolor{gray!20}\textbf{10.89} \\
\bottomrule
\end{tabular}
}
\vspace{-2mm}
\caption{
One-shot pattern detection results on the RPINE-edgeless dataset.
}
\label{tab:RPINE-edgeless}
\vspace{-4mm}
\end{table}

\section{Future work}
\paragraph{Rotation invariance.}
Although \ours effectively handles scale variations, achieving rotation invariance remains a challenging problem, as observed in prior approaches as well~\cite{pseco, geco}.
This limitation could be further mitigated by incorporating rotation-invariant data augmentation or adopting rotation-equivariant architectures~\cite{cohen2016steerable, qureshi2023e2cnn}.

\paragraph{Applications.}
The proposed template-matching based detection framework is potentially useful for detecting low-semantic, user-defined patterns.
One interdisciplinary application is flow cytometry~\cite{givan2011flow, picot2012flow}, which analyzes the physical and chemical characteristics of cell or particle populations~\cite{chen2013flexible, lucas2021locating}, such as in cell counting tasks.
Since repetitive patterns are a fundamental component of many natural and artificial structures~\cite{lee2024diamidocarbene, furukawa2013chemistry}, the framework could be extended to broader applications in real-world settings, such as agricultural or industrial vision.
A thorough investigation of these directions is beyond the scope of this study and is left for future work.

\end{document}